\documentclass[sigconf, nonacm]{acmart}
\usepackage{booktabs}  
\usepackage{threeparttable}
\usepackage{multirow}
\usepackage{enumitem}
\usepackage{subfigure}
\usepackage{graphicx}
\usepackage{xcolor}
\usepackage{algorithmicx,algorithm}
\usepackage{bm}
\usepackage{bbm}
\usepackage{adjustbox}
\usepackage{array}
\usepackage{amsmath}
\usepackage{float}
\usepackage{makecell,rotating,diagbox}
\newcommand\vldbdoi{XX.XX/XXX.XX}
\newcommand\vldbpages{XXX-XXX}
\newcommand\vldbvolume{18}
\newcommand\vldbissue{2}
\newcommand\vldbyear{2024}
\newcommand\vldbauthors{\authors}
\newcommand\vldbtitle{\shorttitle} 
\newcommand\vldbavailabilityurl{https://github.com/ccloud0525/FACTS}
\newcommand\vldbpagestyle{empty}

\begin{document}
\title{Fully Automated Correlated Time Series Forecasting in Minutes}







\author{Xinle Wu$^{1}$, Xingjian Wu$^{2}$, Dalin Zhang$^{1}$, Miao Zhang$^{3}$, Chenjuan Guo$^{2}$, Bin Yang$^{2\ast}$, Christian S. Jensen$^{1}$}
\affiliation{%
  \institution{$^{1}$Aalborg University, Denmark \hspace{5pt} $^{2}$East China Normal University, China \\
  $^{3}$Harbin Institute of Technology, Shenzhen, China
}
  \institution{$^{1}$\{xinlewu, dalinz, csj\}@cs.aau.dk \hspace{5pt} $^{2}$\{xjwu, cjguo, byang \}@dase.ecnu.edu.cn \hspace{5pt} $^{3}$\{zhangmiao@hit.edu.cn\}  }
}

\begin{abstract}
Societal and industrial infrastructures and systems increasingly leverage {sensors} that emit correlated time series. Forecasting of future values of such time series based on recorded historical values has important benefits. Automatically designed models achieve higher accuracy than manually designed models. Given a forecasting task, which includes a dataset and a forecasting horizon, automated design methods automatically search for an optimal forecasting model for the task in a manually designed search space, and then train the identified model using the dataset to enable the forecasting. Existing automated methods face three challenges. First, the search space is constructed by human experts, rending the methods only semi-automated and yielding search spaces prone to subjective biases. Second, it is time consuming to search for an optimal model. Third, training the identified model for a new task is also costly. These challenges limit the practicability of automated methods in real-world settings. To contend with the challenges, we propose a fully automated and highly efficient correlated time series forecasting framework where the search and training can be done in minutes. The framework includes a data-driven, iterative strategy to automatically prune a large search space to obtain a high-quality search space for a new forecasting task. It includes a zero-shot search strategy to efficiently identify the optimal model in the customized search space. And it includes a fast parameter adaptation strategy to accelerate the training of the identified model. Experiments on seven benchmark datasets offer evidence that the framework is capable of state-of-the-art accuracy and is much more efficient than existing methods.

\end{abstract}

\renewcommand*{\authors}{Xinle Wu, Xingjian Wu, Bin Yang, Dalin Zhang, Miao Zhang, Chenjuan Guo, Christian S. Jensen}

\maketitle

\pagestyle{\vldbpagestyle}
\begingroup\small\noindent\raggedright\textbf{PVLDB Reference Format:}\\
\vldbauthors. \vldbtitle. PVLDB, \vldbvolume(\vldbissue): \vldbpages, \vldbyear.\\
\href{https://doi.org/\vldbdoi}{doi:\vldbdoi}
\endgroup
\begingroup
\renewcommand\thefootnote{}\footnote{\noindent
$^*$: Corresponding author. \\
This work is licensed under the Creative Commons BY-NC-ND 4.0 International License. Visit \url{https://creativecommons.org/licenses/by-nc-nd/4.0/} to view a copy of this license. For any use beyond those covered by this license, obtain permission by emailing \href{mailto:info@vldb.org}{info@vldb.org}. Copyright is held by the owner/author(s). Publication rights licensed to the VLDB Endowment. \\
\raggedright Proceedings of the VLDB Endowment, Vol. \vldbvolume, No. \vldbissue\ %
ISSN 2150-8097. \\
\href{https://doi.org/\vldbdoi}{doi:\vldbdoi} \\
}\addtocounter{footnote}{-1}\endgroup

\ifdefempty{\vldbavailabilityurl}{}{
\begingroup\small\noindent\raggedright\textbf{PVLDB Artifact Availability:}\\
The source code, data, and/or other artifacts have been made available at \url{\vldbavailabilityurl}.
\endgroup
}

\section{Introduction}
Many important societal and industrial infrastructures, including intelligent transportation systems, power grids, patient monitoring systems, and industrial control systems~\cite{DBLP:journals/tkde/YangGY22,DBLP:journals/pvldb/PedersenYJ20,rajkumar2010cyber,DBLP:conf/icde/KieuYGJZHZ22,davidsigmod,DBLP:conf/icde/KieuYGCZSJ22,DBLP:conf/icde/YangGHYTJ22,kaivldb24,yunyaovldb24,haoicde24,David2024Qcore,miao2024condensation,xu2024pefad}, involve {sensors} that record values that vary over time, resulting in multiple time series that are often correlated, known as correlated time series (CTS). Forecasting future CTS values based on historical values has important applications~\cite{DBLP:conf/ijcai/WuPLJZ19,DBLP:journals/vldb/GuoYHJC20,davidpvldb,DBLP:journals/tkde/YuHZGYL23}. For example, forecasting the future electricity consumption of users in an area based on their historical electricity consumption can help balance supply and demand in a power grid.

Methods employing deep learning achieve state-of-the-art performance at CTS forecasting. Most of their model architectures are designed manually by human experts~\cite{lai2018modeling,shih2019temporal,DBLP:conf/ijcai/WuPLJZ19,bai2020adaptive,wu2020connecting,razvanicde2021,MileTS,wang2022multivariate,wang2020multi,DBLP:conf/ijcai/CirsteaG0KDP22,DBLP:conf/icde/CirsteaYGKP22,DBLP:journals/tkde/JinZLCYP23,DBLP:journals/pvldb/QiuHZWDZGZJSY24,kieu2024Team,cheng2024memfromer,miao2022mba}. The core components of such models are Spatio-Temporal blocks (ST-blocks), which are constructed by Spatial/Temporal (S/T) operators, such as convolution, graph convolution, and transformer, and capture both temporal dependencies among historical values and spatial correlations across time series.
Although achieving promising results, even human experts struggle to design optimal ST-blocks for new tasks.

\begin{figure*}[htb]
  \centering
  \includegraphics[width=0.75\linewidth]{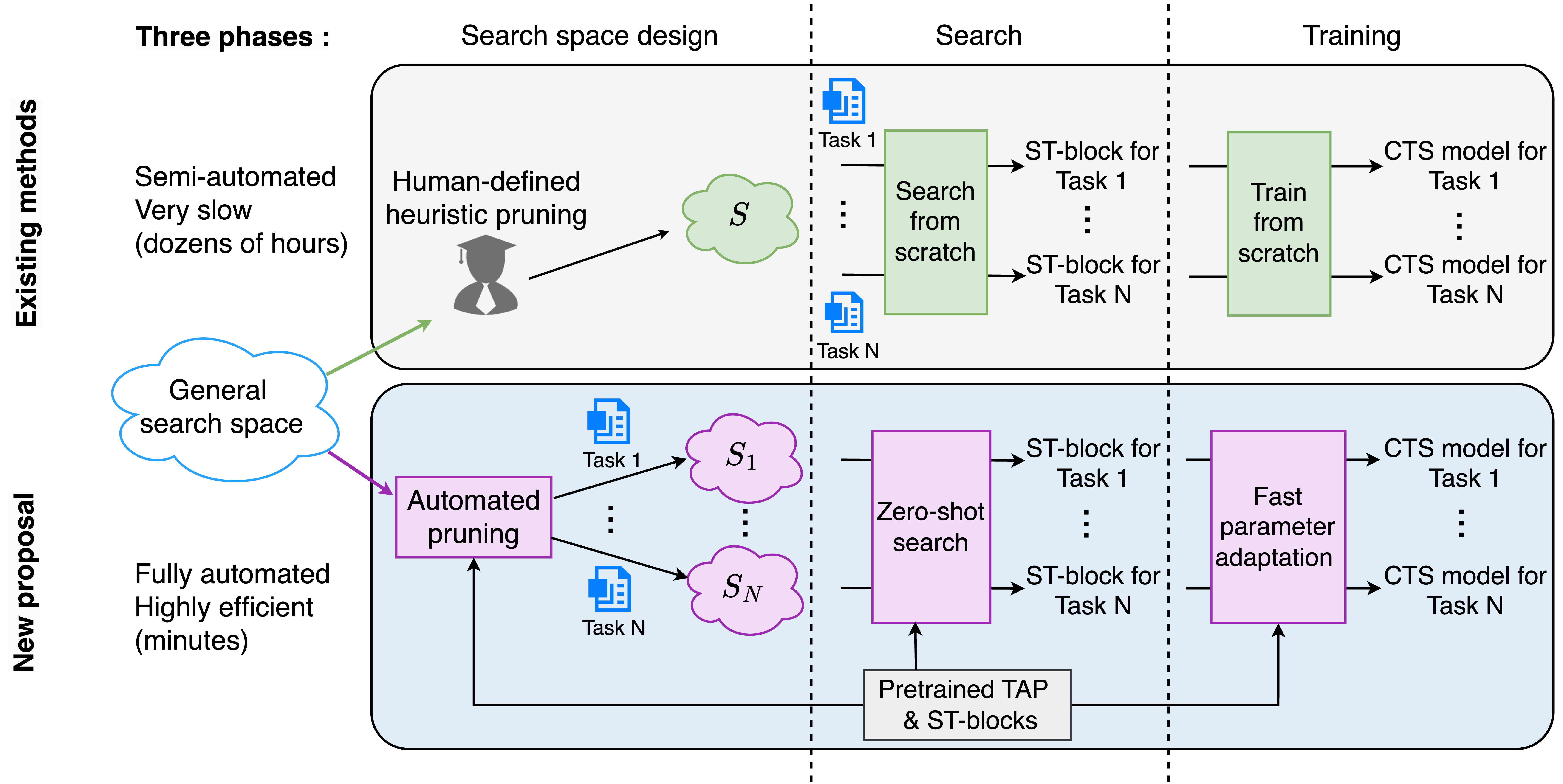}
  \caption{Existing automated methods v.s. the proposed method.
  }
  \label{fig: abstract comparsion}
\end{figure*}

%
As a more promising alternative approach, automated CTS forecasting aims to automatically identify optimal ST-blocks for different tasks and then uses them for forecasting~\cite{pan2021autostg,wu2022autocts,xu2022understanding,wu2023autocts+}.
Figure~\ref{fig: abstract comparsion} illustrates the pipeline of automated methods, which includes three phases: search space design, search for an optimal ST-block, and training the identified ST-block.
A search space is first designed from S/T operators commonly used in existing manually designed models.
These S/T operators are then combined using topological connection rules to obtain a set of ST-blocks that then form the search space.
Next, search strategies, such as gradient-based, comparator-based, or random search, are applied to the search space to find an optimal ST-block for a given task.
Finally, the identified ST-block is trained to enable the forecasting task.
Despite achieving better performance than manually designed models, this approach still suffers from three major limitations that makes it challenging to use in practice.

\noindent \textbf{(1) Manually designed search space.}
The search space is still designed manually, which may yield suboptimal performance and also violates the goal of AutoML, namely to automate the entire process~\cite{DBLP:conf/iclr/ZophL17,he2021automl,ren2021comprehensive,liu2021survey,lee2021rapid}.
Unlike search spaces constructed with a few homogeneous operators in computer vision and natural language processing, the many heterogeneous S/T operators and topological connections possible when forming ST-blocks yield a \textit{general search space} that is difficult to explore.

%
%
%
The state-of-the-art is to prune manually the general search space into a smaller, easier-to-explore search space. However, the pruned search spaces may be suboptimal for unseen tasks because of the manual pruning that relies on heuristic rules and statistical results from a few seen tasks.
\noindent \textbf{(2) High search cost.}
Existing automated methods employ search strategies such as gradient-based~\cite{li2020autost,li2022autost,wu2022autocts}, comparator-based~\cite{wu2023autocts+,chen2021contrastive}, or random search~\cite{xu2022understanding} to explore a search space to find the optimal ST-block, which are all very time-consuming.
Gradient-based methods train a supernet that represents the search space, which is much larger and harder to train than a single ST-block.
Comparator-based methods train a large number of ST-blocks and obtain their validation accuracy to train a comparator.
Random search methods sample a large number of ST-blocks and train them to obtain the optimal one.
As a result, existing automated frameworks spending dozens of GPU hours searching for a {high-performance} ST-block.
This high search cost makes these methods unattractive in real-world scenarios. 


\noindent \textbf{(3) High training cost.}
Having found a {high-performance} ST-block, existing methods train it from scratch on unseen CTS forecasting tasks. Thus usually takes hours~\cite{li2020autost,li2022autost,xu2022understanding}, depending on the scale of the tasks.


To address the above limitations, we propose FACTS, a Fully Automated and highly efficient CTS forecasting framework.
(1) We propose an \textbf{automated pruning strategy} to generate a high-quality search space for unseen forecasting tasks.
Specifically, we construct a general search space using S/T operators and topological rules commonly used in existing CTS forecasting models. This search space is expected to contain optimal ST-blocks on unseen tasks.
We then propose a strategy that partitions the general search space into disjoint subspaces that are distinguishable in terms of quality.
Next, we iteratively prune the search space by removing low-quality subspaces in multiple passes on specific tasks, thereby generating customized high-quality search spaces for different unseen tasks, as shown in Figure~\ref{fig: abstract comparsion} (lower left).

(2) We propose a \textbf{zero-shot search} strategy that can find the optimal ST-block for unseen CTS forecasting tasks in minutes.
Specifically, we build a task-aware architecture predictor (TAP) that takes as input the architecture of an ST-block and the task feature of a task and then predicts the accuracy of the ST-block on the task.
In the pretraining phase, we collect training samples from numerous and diverse CTS forecasting tasks to pretrain TAP, enabling it to predict the accuracy of ST-blocks on an unseen task without having to be trained on that task.
In the zero-shot search phase, we first employ the pretrained TAP to assist search space pruning on the target task, then traverse the ST-blocks in the pruned search space and pick the one with the highest prediction accuracy for deployment.
Figure~\ref{fig: abstract comparsion} (lower middle) shows that FACTS searches for a {high-performance} ST-block for each unseen task.
Since the search on unseen tasks does not involve model training, it can be completed in minutes, which addresses the second limitation.

%
(3) We propose a \textbf{fast parameter adaptation} strategy to accelerate the training of identified ST-blocks. Specifically, we introduce learnable coefficients to linearly combine the parameter weights of pretrained ST-blocks and use these as initial weights in the identified ST-block. This provides a good optimization starting point, allowing the training on the target task to complete quickly, thereby addressing the third limitation.
Figure~\ref{fig: abstract comparsion} (lower right) shows that FACTS inherits the parameter weights from pretrained ST-blocks to train identified ST-blocks, resulting in trained CTS models for forecasting.

%
Our contributions are summarized as the follows.
\begin{itemize}
    \item [(1)]
    We propose FACTS, a fully automated CTS forecasting framework. In particular, we propose an automated search space pruning strategy to automatically generate high-quality search spaces for unseen CTS forecasting tasks.
    \item [(2)]
    We propose a zero-shot search strategy to search for an optimal ST-block on arbitrary unseen CTS forecasting tasks in minutes.
    \item [(3)]
    We propose a fast parameter adaptation strategy to accelerate the training of identified ST-blocks, reducing the training time by up to 66\% on unseen CTS forecasting tasks.
    \item [(4)]
    Extensive experiments on seven benchmark datasets show that the proposed framework is capable of state-of-the-art forecast accuracy while taking less time than existing manual and automated methods.
\end{itemize}

%
%

\section{Preliminaries}
\label{Sec: Prelim}
\subsection{Problem Setting}

\noindent\textbf{Correlated Time Series (CTS).}
A correlated time series (CTS) $\bm{\mathcal{X}}$ consists of $N$ times series that each contains $T$ timestamps and has an $F$-dimensional feature vector for each timestamp. Thus, $\bm{\mathcal{X}} \in \mathbb{R}^{N\times T\times F}$. Time series are correlated when the values in each time series not only depend on its historical values but also depend on values in other time series. Correlations between time series can be captured using a graph $G=(V,E,A)$. where $V$ is a set of vertices representing time series, $E$ is a set of edges,
representing the correlations between two time series,
such correlations are often derived from the physical distances between the sensors that produce the time series, but they can also be learned adaptively. A graph $G$ can be captured using an adjacency matrix $A$.

\noindent\textbf{Correlated Time Series Forecasting.}
We consider multi-step and single-step CTS forecasting, both of which have important applications. Given the feature values of a CTS $\bm{\mathcal{X}}$ in the past $P$ time steps, the goal of multi-step forecasting is to predict the feature values in $Q$ ($Q$ > 1) future time steps, formulated as follows.
\begin{align}
{\{\bm{\hat{X}}_{t+P+1}, \bm{\hat{X}}_{t+P+2}, ...,\bm{\hat{X}}_{t+P+Q}\}}=\mathcal{F}(\bm{X}_{t+1}, \bm{X}_{t+2}, ..., \bm{X}_{t+P};G),
\label{form: multi}
\end{align}
where $\bm{X}_{t} \in \mathbb{R}^{N\times F}$ denotes the feature values of a CTS at time step $t$, $\bm{\hat{X}}$ denotes forecasted feature values, and $\mathcal{F}$ is a forecasting model.
The goal of single-step forecasting is to forecast the feature values in the $Q$-th ($Q \geq 1$) future time step, formulated as follows.
\begin{align}
\bm{\hat{X}}_{t+P+Q}=\mathcal{F}(\bm{X}_{t+1}, \bm{X}_{t+2}, \ldots, \bm{X}_{t+P};G),
\label{form: single}
\end{align}


\noindent\textbf{CTS Forecasting Task.}
We define a CTS forecasting task as $\mathcal{T} = (\mathcal{D},P,Q,tag)$, where $\mathcal{D}$ represents a CTS dataset, $P$ and $Q$ represent the input and forecasting lengths, respectively, and $tag$ indicates whether the CTS forecasting task is single-step or multi-step.

\subsection{Existing Automated Forecasting Methods}

\begin{figure}[htb]
  \centering
  \includegraphics[width=0.77\linewidth]{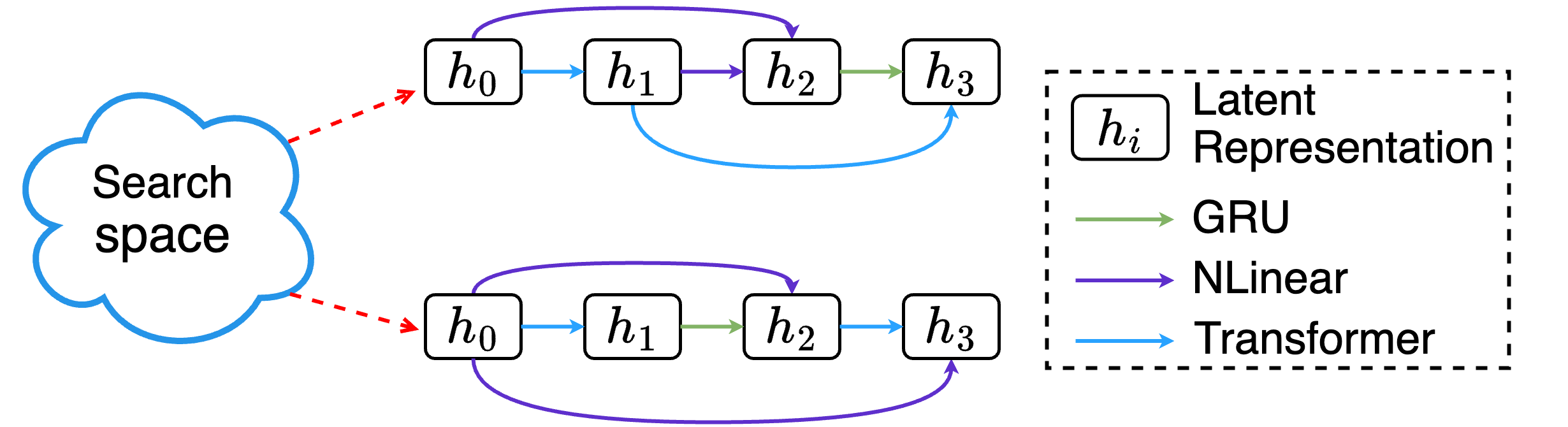}
  \caption{Example of a search space and ST-blocks. 
  }
  \label{fig: space}
\end{figure}

%
Existing methods focus on the automated design of ST-blocks, which form the backbone of CTS forecasting models.
Specifically, automated CTS forecasting includes three phases: search space design, search for an optimal ST-block, and training of the identified ST-block, as shown in Figure~\ref{fig: abstract comparsion}.
First, a search space is designed, which is a set of ST-blocks composed of \textbf{S/T operators} and \textbf{topological connection rules} used to assemble the operators.
Figure~\ref{fig: space} shows an example of a search space and two ST-blocks in it, which feature different S/T operator combinations and topologies.
Next, they employ search strategies such as gradient-based, comparator-based to explore the search space for the optimal ST-block.
Finally, they train the identified ST-block for CTS forecasting.

\subsubsection{Search Space Design.}
\label{pre:space}
{Existing automated CTS forecasting methods employ commonly used S/T-operators to build ST-blocks. We use $O$ to denote the S/T-operator set. It includes $9$ T-operators: 1D-CNN~\cite{schmidhuber2015deep}, LSTM~\cite{hochreiter1997long}, GRU~\cite{cho2014learning}, NLinear~\cite{zeng2023transformers}, GDCC~\cite{DBLP:conf/ijcai/WuPLJZ19}, Inception~\cite{wu2020connecting}, Transformer~\cite{vaswani2017attention}, Informer~\cite{zhou2021informer}, and Convformer~\cite{gu2023convformer}. Further, it includes $5$ S-operators: DGCN~\cite{DBLP:conf/ijcai/WuPLJZ19}, Mix-hop~\cite{wu2020connecting}, Spatial-Transformer~\cite{vaswani2017attention}, Spatial-Informer~\cite{zhou2021informer}, Masked-Transformer~\cite{jiang2023pdformer}. Finally, it includes a skip operator. Thus, $|O|=15$.}
These operators are used to design a \textbf{general search space} that is expected to contain optimal ST-blocks for arbitrary unseen CTS forecasting tasks, but is too large to explore to find the optimal ST-block.
Existing automated methods then prune the general search space manually, in two ways: by reducing the S/T operator set~\cite{pan2021autostg,li2022autost,li2020autost}or by removing bad ST-blocks based on sampling~\cite{xu2022understanding,radosavovic2020designing}.

Methods in the first category~\cite{pan2021autostg,li2022autost,wu2022autocts} compare the accuracy and efficiency of individual operators and select a subset of high-performance operators from the full operator set $O$, thereby pruning the general search space. However, the pruned search space may be suboptimal. First, it compares the accuracy of S/T-operators on a few seen CTS forecasting tasks, which may not generalize to unseen tasks.
For example, as shown in Table~\ref{tab: singleOP}, NLinear performs better than a 1D-CNN on the ETTH dataset, but worse than the 1D-CNN on the PEMS04 and PEMS08 datasets. Therefore, it is unwise to remove the NLinear operator from the search space based on the results on PEMS04 and PEMS08, as it may exhibit high performance on other datasets.
Second, S/T operators with high performance in isolation may not form high-performance ST-blocks.
For example, as shown in Table~\ref{tab: singleOP}, Convformer (ConvF) has higher accuracy than Informer (INF) on multiple datasets, but when we replace INF with ConvF in multiple ST-blocks, the accuracy decreases---see ST-block (INF) and ST-block (CovF) in Table~\ref{tab: singleOP} for an example, where ST-block (CovF) is a variant of ST-block (INF) with CovF replacing INF.


Methods in the second category~\cite{xu2022understanding,radosavovic2020designing} adopt heuristic pruning strategies to divide architecture designs into disjoint angles and remove low-performance designs based on the accuracy of sampled ST-blocks, which is also performed on a few seen tasks and may not generalize to unseen tasks.


%
To sum up, existing methods prune the general search space based on \textbf{human designed heuristics} to generate a task-agnostic and suboptimal search space. Instead, FACTS \textbf{automatically} prunes the general search space for the target task to generate a task-aware search space, which is expected to offer a higher potential for finding high-performance ST-blocks for the specific target task.


\begin{table}[h]
\small
    \centering
    \caption{MAE of S/T operators.}
    \resizebox{\linewidth}{!}{
    \begin{tabular}{c|c|c|c|c|c|c}
        \hline
        &1D-CNN&NLinear&ConvF&InF&ST-block (INF)&ST-block (ConvF) \cr
        \hline
    ETTh1&0.535&0.377&0.514 &0.635 &0.203  &0.224  \cr
    PEMS04&31.556&36.484&26.092&29.084&18.954&19.104 \cr
    PEMS08&26.257&30.448&20.709&23.693&14.694&14.870  \cr
    \hline
    \end{tabular}
    }
    \label{tab: singleOP}
\end{table}

\subsubsection{Search Strategy.}
Existing search strategies include gradient-based~\cite{li2020autost,li2022autost,pan2021autostg}, comparator-based~\cite{wu2023autocts+,chen2021contrastive}, and random search~\cite{xu2022understanding}.
The gradient-based search strategy models the search space as a supernet, which is time-consuming to train.
The comparator-based search strategy trains a comparator to identify the better of two ST-blocks, which requires training a large number of ST-blocks and is therefore also costly.
The random search strategy samples a number of ST-blocks at random and trains them to obtain the one with the highest accuracy, thus having the highest search cost.

\subsubsection{Model Training.}
Existing automated methods train an identified ST-block from scratch~\cite{li2020autost,wu2022autocts,wu2023autocts+,pan2021autostg,xu2022understanding,dong2020autohas}, which typically takes hours, depending on the scale of the forecasting task.

\section{Methodology}
\begin{figure*}[t]
  \centering
  \includegraphics[width=0.77\linewidth]{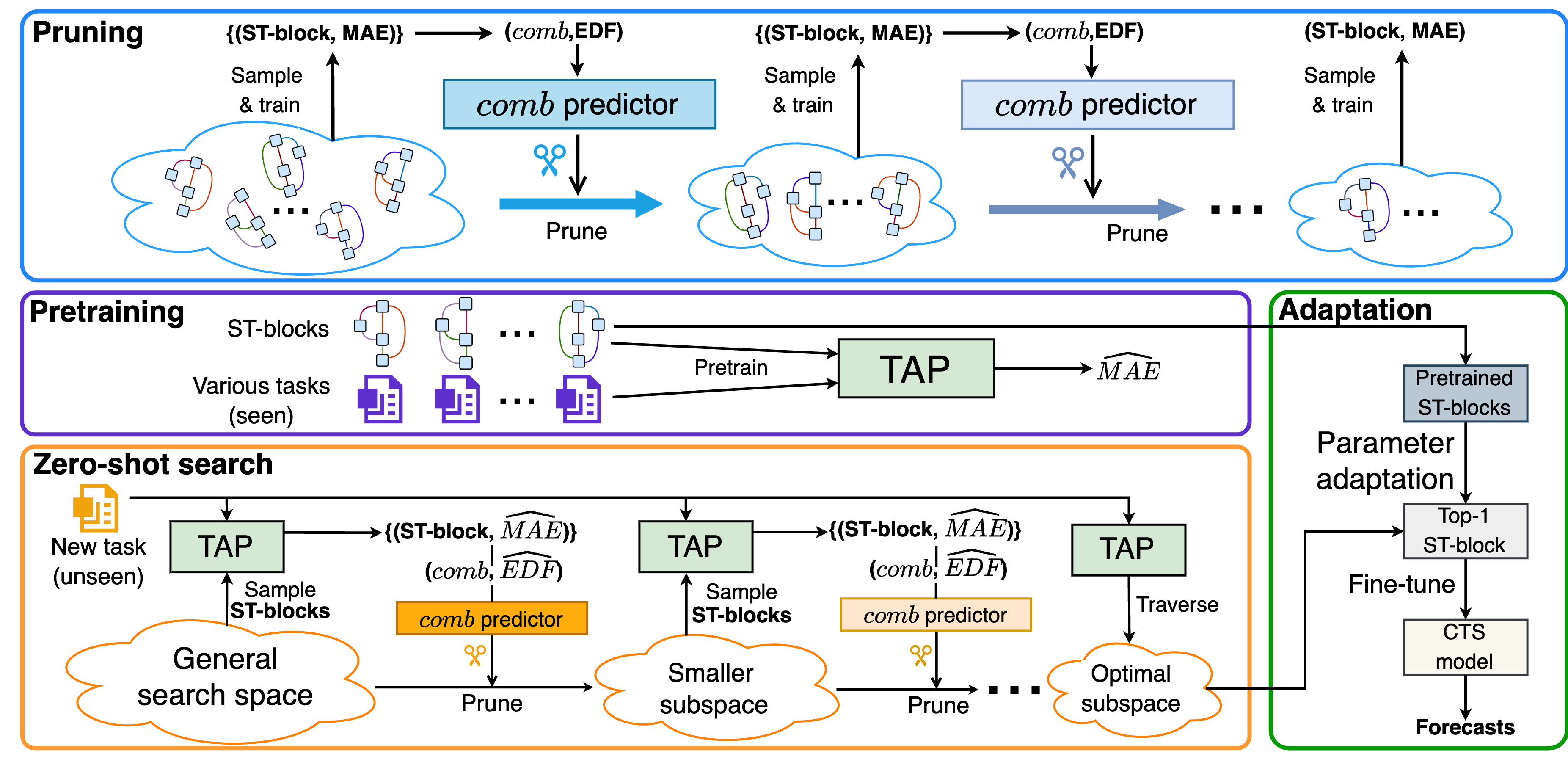}
  \caption{The FACTS framework.
  }
  \label{fig: frame}
\end{figure*}

The proposed framework aims to find an ST-block that enables optimal prediction accuracy on unseen CTS forecasting tasks and to use it to obtain results in minutes. Figure~\ref{fig: frame} illustrates the framework.
It includes an \textbf{automated pruning strategy} (Section~\ref{ssec: prune}) to prune the general search space to obtain a high-quality and relatively small search space suitable for unseen CTS forecasting tasks.
Specifically, we first partition the general search space into disjoint subspaces, named $combs$. 
{We define the quality of a $comb$ to be a function of the performance of all ST-blocks in it and measure the quality using the error empirical distribution function (EDF)~\cite{radosavovic2020designing}.}
We then train a $comb$ predictor to predict the EDF of $combs$ and prune the search space by iteratively removing $combs$ with low EDF.
%
%

%
To accelerate the search for optimal ST-blocks, the framework includes a highly efficient \textbf{zero-shot search strategy} (Section~\ref{ssec: pretrain}), which consists of pretraining and zero-shot search phases. Pretraining is a one-time task, and zero-shot search can be completed on unseen tasks in minutes.
In the pretraining phase, we perform search space pruning on numerous and diverse CTS forecasting tasks, during which we collect training samples to pretrain a task-aware architecture predictor (TAP).
In the zero-shot search phase, we first perform automated search space pruning on an unseen task, where we use the pseudo-EDF generated by the pretrained TAP to replace the real EDF.
With the pruning completed, we use the pretrained TAP to predict the accuracy of all ST-blocks in the pruned search space on the unseen task and return the ST-block with the best predicted performance as the optimal ST-block.
%

%
Finally, we train the identified ST-block on the unseen CTS forecasting task to enable forecasting. 
To accelerate the training, we propose a \textbf{fast parameter adaptation strategy} (Section~\ref{ssec: adapt}) that introduces learnable coefficients to linearly combine the parameter weights of pretrained ST-blocks and serve as the initial parameter weights of the identified ST-block, which accelerates convergence and reduces the training time by up to 66\%.

\subsection{Automated Search Space Pruning}
\label{ssec: prune}

We propose an automated strategy to prune the general search space into a high-quality and relatively small search space for a specific 
CTS forecasting task.
To achieve this, we first partition the general search space into subspaces that are distinguishable in quality and then remove lower-quality subspaces iteratively.
Next, we introduce the search space partitioning method, the EDF metric used to evaluate the quality of search spaces, and the iterative search strategy.



\subsubsection{Search Space Partitioning}
\label{ssec: partition}
Before introducing the search space partitioning method, {we first introduce the general search space and define the notion of a $comb$.}

{The \textbf{general search space} is a collection of a large number of ST-blocks, each of which is constructed by S/T-operators combined using topological connection rules.
%
We use the S/T-operators introduced in Section~\ref{pre:space} and follow commonly used topological connection rules~\cite{liu2018darts,pan2021autostg,wu2022autocts} to assemble them into ST-blocks.
}


%
\begin{definition}
{
A $comb$ is a search space containing ST-blocks with the same S/T-operator combinations. It is represented as $comb=[n_{o_1}, n_{o_2}, ..., n_{o_L}]$, where $n_{o_i}$ is the number of occurrences of S/T-operator ${o_i}$ in each ST-block in the $comb$ and $L$ is the number of S/T-operators used for constructing the search space. Note that $0 \leq n_{o_i} \leq n_o$, where $n_o$ is the number of S/T-operators contained in an ST-block.}
\end{definition}
%
For example, consider a search space constructed with S/T operators $\mathit{skip}$, $\mathit{CNN}$, $\mathit{GCN}$, Informer-Temporal ($\mathit{INF}$-$T$), Informer-Spatial($\mathit{INF}$-$S$), and $\mathit{GRU}$. Then $comb=[0, 2, 2, 0, 1, 2]$ is a search space in which the ST-blocks contain 0 $\mathit{skip}$, 2 $\mathit{CNN}$, 2 $\mathit{GCN}$, 0 $\mathit{INF}$-$T$, 1 $\mathit{INF}$-$S$ and 2 $\mathit{GRU}$, but with different topological connections.

\begin{table}[h]
    \small
    \centering
    \caption{Comparison of different $combs$.}
    \resizebox{0.95\linewidth}{!}{
    \begin{tabular}{c|c|c|c|c|c|c|c}
    \toprule  &{$\mathit{skip}$}&{$\mathit{CNN}$}&{$\mathit{GCN}$}&{$\mathit{INF}$-$T$}&{$\mathit{INF}$-$S$}&{$\mathit{GRU}$}&{$EDF$} \cr
    \hline
    {$comb_1$}&{0}&{2}&{2}&{0}&{1}&{2}&{0.16} \cr
    \hline
    {$comb_2$}&{1}&{1}&{2}&{1}&{1}&{1}&{0.22} \cr
    \hline
    {$comb_3$}&{1}&{2}&{1}&{2}&{0}&{1}&{0.54} \cr
    \hline
    {$comb_4$}&{1}&{2}&{1}&{2}&{1}&{0}&{0.48} \cr
    \hline
    {$comb_5$}&{1}&{2}&{2}&{2}&{0}&{0}&{0.69} \cr
    \bottomrule
    \end{tabular}
     }
    \label{tab: comb}
\end{table}

%
The partitioning method is based on the observation that the combination of S/T-operators largely determines the accuracy of the ST-blocks they compose, resulting in different $combs$ being distinguishable in quality, where the quality of a $comb$ refers to the overall performance of the ST-blocks in it and is measured using EDF (see Section~\ref{ssec: EDF}).
%
To illustrate this phenomenon, we select 5 $combs$ and randomly sample 200 ST-blocks in each and calculate its EDF.
As we can see in Table~\ref{tab: comb}, the EDF of the different $combs$ are clearly distinguishable. In particular, $comb_1$ and $comb_2$ have significantly lower quality, indicating that we can remove the two $combs$ to prune the search space, resulting in a smaller and higher-quality pruned search space.
We thus propose to partition a search space into $combs$ and prune it by removing low-quality $combs$.
\begin{figure}[b]
  \centering
  \includegraphics[width=0.70\linewidth, trim=90 70 70 70,clip]{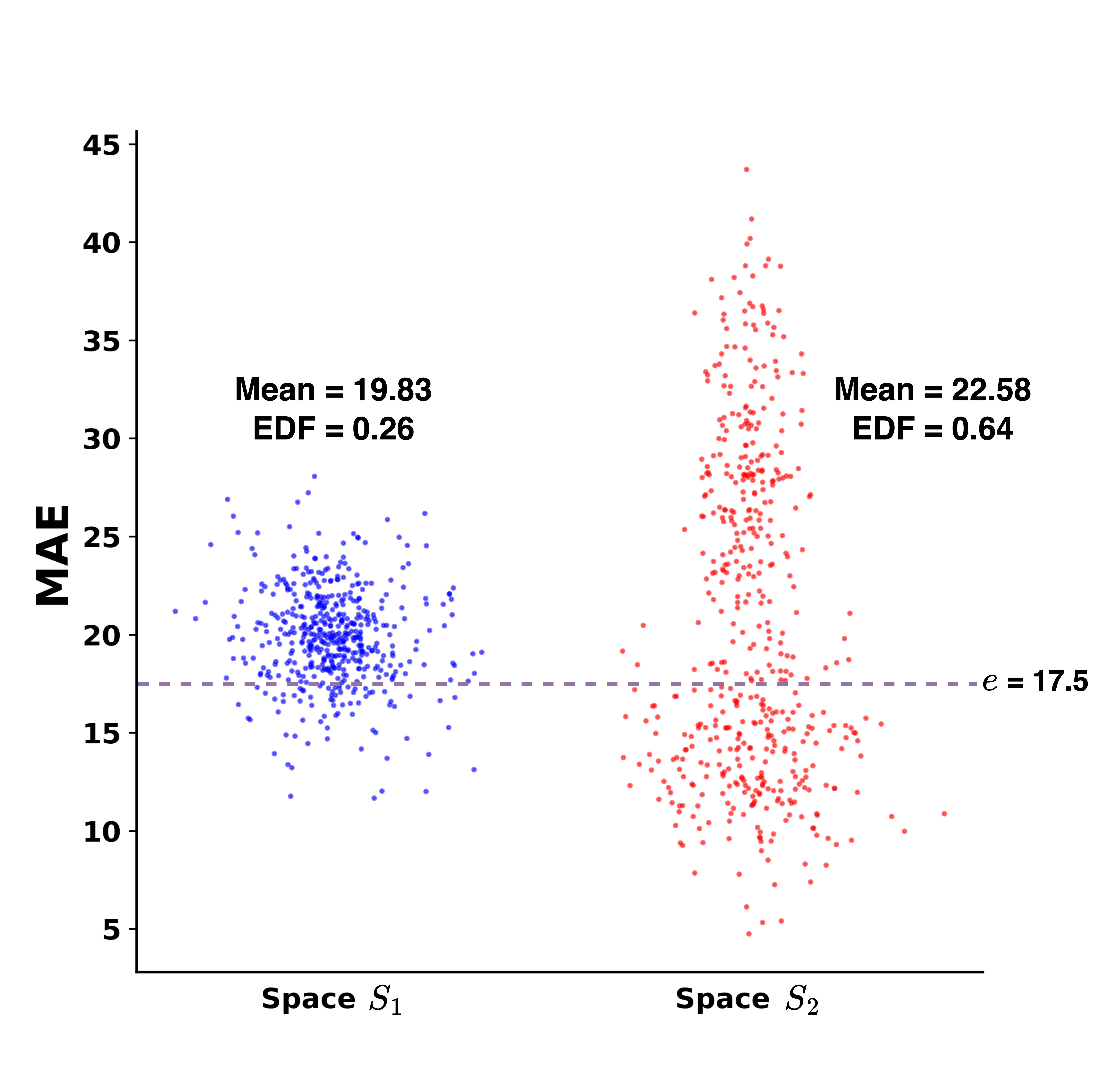}
  \caption{Search space quality metrics. 
  }
  \label{fig: EDF}
\end{figure}

\subsubsection{Search Space Quality Evaluation}
\label{ssec: EDF}


{The quality of a search space $\mathcal{S}$ aims to capture the overall performance of the ST-blocks in it.
We thus define the quality of $\mathcal{S}$ as a function of the performance of all ST-blocks in it.
\begin{align}
quality(\mathcal{S}) = \mathit{g}(\mathit{p}(b_1), \mathit{p}(b_2), ..., \mathit{p}(b_{n})),
\label{form: quality}
\end{align}
where $\mathit{p}(b_i)$ is the performance of ST-block $b_i$, which is evaluated using validation accuracy, and $\mathit{g}$ is a function that aggregates the performance of the constituent blocks.} 

Since it is impractical to train all ST-blocks to obtain their accuracy, we use uniform sampling to approximate.
A simple metric to evaluate the quality of a search space is the average accuracy of the sampled ST-blocks, but this does not accurately reflect the quality of the search space.
To illustrate this, we select $combs$ $S_1$ and $S_2$ and randomly sample 200 ST-blocks in each and train these ST-blocks to obtain their validation accuracy. We show in Figure~\ref{fig: EDF} the MAE of ST-blocks sampled from the two $combs$, where a lower MAE corresponds to a higher accuracy.
We see that $S_2$ has many more high-accuracy ST-blocks than $S_1$, 
indicating that $S_2$ is of higher quality than $S_1$, as our goal is to find a single optimal ST-block in the search space.
However, the average accuracy of sampled ST-blocks in $S_1$ is higher than that in $S_2$. Thus, the average accuracy of sampled ST-blocks fails to reflect the quality of a search space.

Instead, we use the error empirical distribution function (EDF) metric~\cite{radosavovic2020designing} to quantify the quality of a search space.
EDF captures the quality of a search space by randomly sampling a set of ST-blocks and calculating the proportion of ST-blocks whose MAE error is below a given threshold $e$:
\begin{align}
F(e) = \frac{1}{n_e}\sum\limits_{j=1}^{n_e}\mathbf{1}\left[e_j<e\right],
\label{eq: EDF}
\end{align}
where $n_e$ is the number of sampled ST-blocks and $e_j$ is the validation MAE error of the $j$-th ST-block.
In the context of CTS forecasting, the intuition is that comparing the proportion of high-accuracy ST-blocks is more robust than using the average metric, as it reduces the impact of outliers.
In the case of Figure~\ref{fig: EDF}, the EDF of $S_2$ is significantly higher than that of $S_1$, reflecting the true quality difference between the two search spaces.

With the $comb$-based search space partitioning method, and using the EDF metric to evaluate the quality of a $comb$, a naive idea to reduce the search space is to calculate the EDF of all $combs$ and remove low-quality $combs$. However, the number of $combs$ is huge, and uniformly sampling ST-blocks in each $comb$ to calculate its EDF is too costly.
Instead, we only sample $K$ $combs$ and calculate their EDF values, and use the collected ($comb$, EDF) samples to train a $comb$ predictor to predict the EDF of all remaining $combs$. For efficiency, we employ a {Gradient Boosting Decision Tree (GBDT)~\cite{friedman2001greedy}} model as the $comb$ predictor.


\subsubsection{Iterative Search Space Pruning}
\label{iter}
\begin{table}[b]
    \centering
    \caption{EDF values of different combinations.}
    \resizebox{0.9\linewidth}{!}{
    \begin{tabular}{l|l|l|l|l|l|c|c}
    \toprule  \emph{}&\emph{MAE}&\emph{MAE}&\emph{MAE}&\emph{MAE}&\emph{MAE}&\emph{EDF ($e=16.00$)}&\emph{EDF ($e=14.70$)} \cr
    \hline
    {$comb_1$}&{14.65}&{14.72}&{14.73}&{14.75}&{17.36}&{0.8}&{0.2} \cr
    \hline
    {$comb_2$}&{14.85}&{14.87}&{14.89}&{14.93}&{18.50}&{0.8}&{0.0} \cr
    \hline
    {$comb_3$}&{15.38}&{15.54}&{15.88}&{16.75}&{20.33}&{0.6}&{0.0} \cr
    \bottomrule
    \end{tabular}
    }
    \label{tab: EDF}
\end{table}
Pruning the search space only once does not guarantee a high-quality pruned search space. This is because the threshold $e$ in Equation~\ref{eq: EDF} needs to be set manually, and an inappropriate $e$ may cause EDF to fail to accurately compare the quality of different subspaces.
To illustrate this, we select 3 $combs$ and randomly sample 5 ST-blocks in each. We then train these ST-blocks to obtain their validation MAE error.
As shown in Table~\ref{tab: EDF}, when $e$ is set to the relatively large value of 16.00, we can easily identify $comb_3$ as the worst $comb$, but $comb_1$ and $comb_2$ have the same EDF, although ST-blocks in $comb_1$ perform significantly better than those in $comb_2$. This demonstrates that a large $e$ may cause EDF to be unable to distinguish between good and relatively good $combs$, which may cause us to remove high-quality subspaces, thereby reducing the potential of the pruned search space.
In contrast, when $e$ is set to the relatively small value of 14.70, the EDF labels of both $comb_2$ and $comb_3$ are 0, although the sampled ST-blocks from $comb_2$ have significantly higher accuracy than those from $comb_3$.
Further, since we need to collect ($comb$, EDF) samples to train a $comb$ predictor to predict the EDF of all remaining $combs$, too many 0 labels will cause the training samples to be unbalanced, resulting in an inaccurate $comb$ predictor.

To solve this problem, we iteratively prune the search space, reducing $e$ gradually. The intuition is that a larger $e$ in the initial stage allows us to easily identify and remove bad $combs$. As $e$ decreases, we can gradually distinguish good and relatively good $combs$, and since the quality of the remaining $combs$ gradually becomes higher, there will not be a large number of ($comb$, EDF) samples with EDF values equal to 0.
Specifically, we first divide the general search space into $M_0$ different $combs$.
In the $i$-th pruning stage, we randomly sample $c$ $combs$, and for each $comb$, we randomly sample $r$ ST-blocks to calculate its EDF. Then, we train a $comb$ predictor using the ($comb$, EDF) samples collected from iterations $0$ to $i$ and use it to predict the EDF values of all remaining $combs$. We end by pruning the search space by removing the 50\% of $combs$ with the lowest EDF values.
The pruning ends when the number of remaining $combs$ is below $M$.
We set the initial threshold $e_0$ to be the median accuracy of the ST-blocks sampled in the $0$-th iteration, and we set the final threshold $e$ to be the $\lfloor M/M_0 \rfloor$-th best accuracy of the ST-blocks sampled in the $0$-th iteration. We then reduce the threshold linearly during the iterative pruning.
{The complete pruning procedure is shown in Algorithm~\ref{alg}.}

\begin{algorithm}[!htbp]
\caption{{Search Algorithm}}
\begin{flushleft}
{
{\bf Input:} 
CTS dataset $\mathcal{D}$, general search space $S_0$}

{
{\bf Output:}
a pruned search space $S_{n_p}$
}
\end{flushleft}

\begin{algorithmic}[1]
\State {Split $\mathcal{D}$ into $\mathcal{D}_\mathit{train}$, $\mathcal{D}_\mathit{val}$, and $\mathcal{D}_\mathit{test}$}
\State {Split $\mathcal{S}_0$ into $M_0$ $combs$ and calculate the iteration number $n_p$ with $n_p = \lceil \log_{2}{M_0/M}\rceil$}
\State {Randomly sample $c$ $combs$ to form a $comb$ set $\mathcal{C}$; then randomly sample $r$ ST-blocks from each $comb$ to form an ST-block set $\mathcal{B}$}
\State {Train ST-blocks in $\mathcal{B}$ on $\mathcal{D}_{train}$ and obtain their validation accuracy on $\mathcal{D}_{val}$}
\State {Set the initial threshold $e_0$ to the median accuracy of the ST-blocks in $\mathcal{B}$ and set the final threshold $e$ to the $\lfloor M/M_0 \rfloor$-th best accuracy of the ST-blocks in $\mathcal{B}$}
\State {{\bf for} i = 0, ..., $n_p$-1 {\bf do}}
\State \hspace{0.1in} {Calculate the current threshold with $e_i = e_0 - \frac{e_0-e}{n_p-1}\cdot i$}
\State \hspace{0.1in} {Calculate EDF values for $combs$ in $\mathcal{C}$ using Equation~\ref{eq: EDF}}
\State \hspace{0.1in} {Train a GBDT model with ($comb$, EDF) pairs in $\mathcal{C}$}
\State \hspace{0.1in} {Use the GBDT model to predict EDF values for $combs$ in $S_i$}
\State \hspace{0.1in} {Remove the 50\% of the $combs$ with the lowest EDF values; the remaining $combs$ form a new sub-search space $S_{i+1}$}
\State \hspace{0.1in} {{\bf if} i = $n_p$-1}
\State \hspace{0.2in} {{\bf break}}
\State \hspace{0.1in} {Randomly sample $c$ $combs$ from $S_{i+1}$ to form a $comb$ set $\mathcal{C}_i$, then randomly sample $r$ ST-blocks from each $comb$ in $\mathcal{C}_i$ to form an ST-block set $\mathcal{B}_i$}
\State \hspace{0.1in} {Train ST-blocks in $\mathcal{B}_i$ on $\mathcal{D}_{train}$ and obtain their validation accuracy on $\mathcal{D}_{val}$}
\State \hspace{0.1in} {$\mathcal{C} \leftarrow \mathcal{C} \cup \mathcal{C}_i$}
\State {{\bf end for}}
\State {{\bf return} The search space $S_{n_p}$ formed by the remaining $combs$}
\end{algorithmic}
\label{alg}
\end{algorithm}

\subsection{Zero-Shot Search}
\label{ssec: pretrain}
We pretrain a TAP on numerous and diverse CTS forecasting tasks and then perform zero-shot search on unseen tasks to find optimal ST-blocks in minutes.
We first introduce the structure of the TAP and then describe how to pretrain it and perform zero-shot search.

\subsubsection{Task-aware Architecture Predictor}
\label{sssec: predictor}
\begin{figure}[b]
  \centering
  \includegraphics[width=0.85\linewidth]{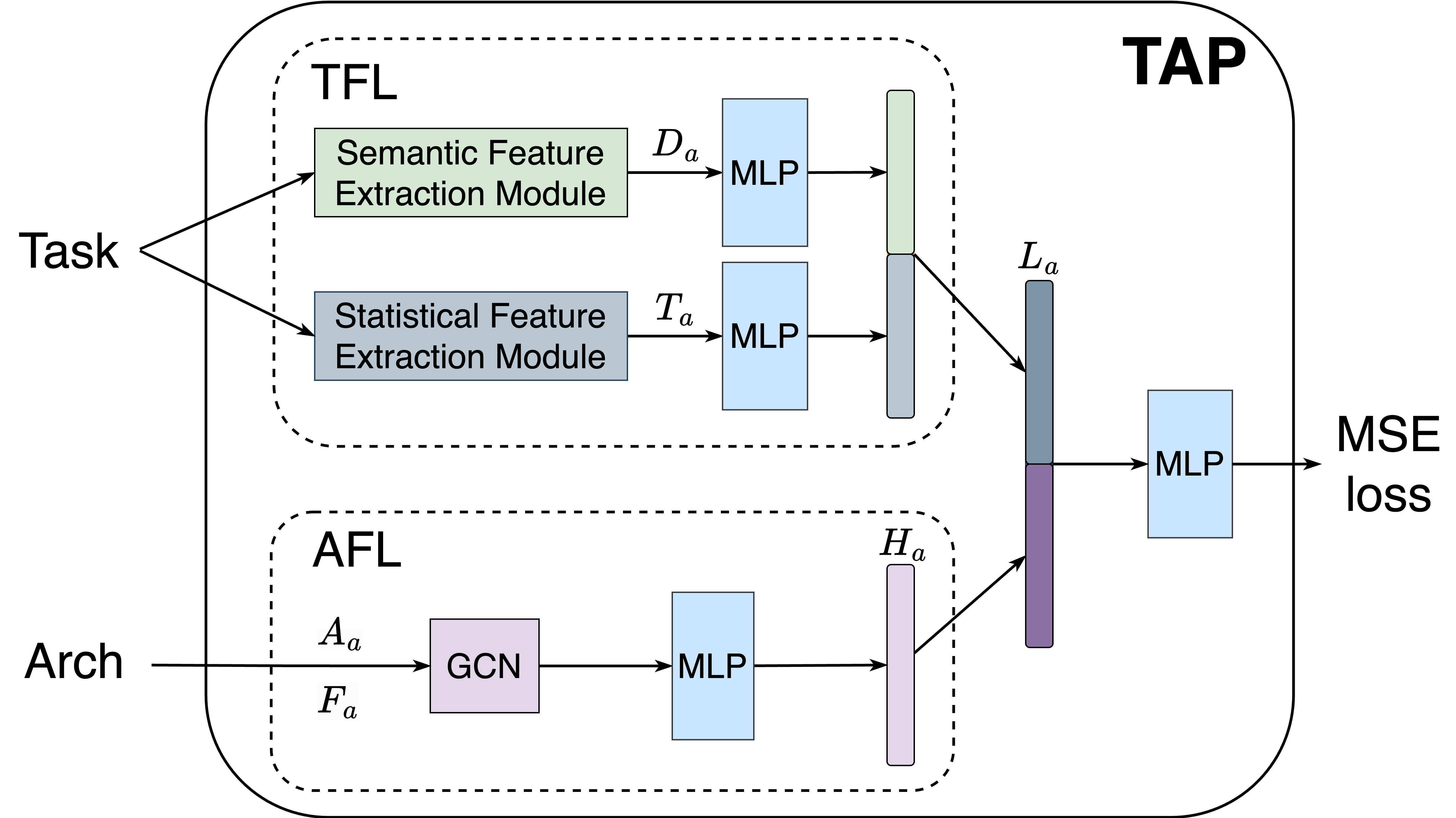}
  \caption{Task-aware Architecture Predictor. 
  }
  \label{fig: TAP}
\end{figure}

We propose a Task-aware Architecture Predictor (TAP) that takes the architecture of an ST-block and a task as input and outputs the prediction accuracy of the ST-block.
Figure~\ref{fig: TAP} shows the structure of the TAP, which consists of an Architecture Feature Learning (AFL) module, a Task Feature Learning (TFL) module, and an MLP regressor.

AFL extracts features of the architecture of an ST-block. We regard the architecture of an ST-block as a directed acyclic graph (DAG) and represent it by an adjacency matrix $A_a$ and a feature matrix $F_a$. We then use a graph convolution network (GCN) followed by a single-layer MLP to encode $A_a$ and $F_a$ as a feature vector $H_a$, formulated as follows.

\begin{align}
H_a = MLP_1(GCN(A_a, F_a))
\end{align}

TFL extracts features of a CTS forecasting task. We consider both semantic features and statistical features to learn an informative vector representation of a task.
We use a 2-layer Set-Transformer~\cite{lee2019set} to extract the semantic feature vector of a task. {The first layer, named \textit{IntraSetPool}, consists of a Set Attention Block (SAB)~\cite{lee2019set} for capturing the relationship between different samples of a single time series and a Multi-head Attention (PMA)~\cite{lee2019set} for pooling the samples into a single representation of the time series.
The second layer, named \textit{InterSetPool}, also consists of an SAB and a PMA, where the former is to capture the relationship between different time series and the latter is to pool these time series into a single representation of the task.
We refer to Set-Transformer~\cite{lee2019set} for the detailed architecture of SAB and PMA.
Formally, we have}
\begin{align}
\{\Tilde{D_i}\} &= \mathit{IntraSetPool}(\{{D_i}\}, W_1) \\
D_a &= \mathit{InterSetPool}(\{\Tilde{D_i}\}, W_2),
\label{eq: set-trans}
\end{align}
where $\{D_i\}$ is a CTS forecasting dataset, $D_a$ is the learned semantic feature vector, and $W_1$ and $W_2$ are learnable parameters of the Set-Transformer.
%
In addition, we use tsfresh~\cite{christ2018time} for extracting statistical features to construct a statistical feature vector $T_a$.
We input the semantic and statistical feature vectors into different single-layer MLPs to achieve feature alignment, and then we concatenate the aligned feature vectors with the feature vector $H_a$ and feed it into a two-layer MLP regressor to predict the accuracy of the ST-block on the CTS forecasting task, which is formulated as follows.
\begin{align}
L_a &= concatenate(MLP_2(D_a), MLP_3(T_a), H_a) \\
output &= MLP_4(L_a)
\end{align}
Finally, we optimize the parameters of TAP using the mean squared error (MSE) loss.

\subsubsection{TAP Pretraining}
\label{sssec: TAP_sample}
We pretrain TAP on numerous and diverse CTS forecasting tasks, enabling it to predict the accuracy of ST-blocks on arbitrary unseen tasks, thereby performing zero-shot search.
The intuition is that we learn a combined embedding space of ST-blocks and tasks through pretraining, as well as a mapping from embeddings in the space to the accuracy of ST-blocks.
When using the pretrained TAP to predict the accuracy of an ST-block on an unseen task, we first use AFL and TFL to obtain their combined embedding, and then we use the MLP regressor in the TAP to map the embedding to the accuracy of the ST-block.

%
We collect training samples of the form ($t$, $b$, $R(t, b)$) to pretrain the TAP, where $R(t, b)$ represents the validation accuracy of ST-block $b$ on task $t$ and is obtained by fully training $b$ on $t$.
Given a fixed training sample budget, a naive idea is to collect training samples by uniformly sampling ST-blocks from the general search space.
However, the general search space is so large that it is difficult to learn the mapping from the input $b$ and $t$ to accuracy with limited training samples.

Considering that our goal is to find the optimal ST-block, we are more concerned with accurate prediction of high-performance ST-blocks than with accurate prediction of low-performance regions in the general search space.
Therefore, we propose to collect training samples iteratively during the search space pruning process.
As the search space is pruned iteratively, training samples are collected gradually from the pruned high-quality search space, so the TAP trained with these samples has more accurate prediction capabilities for high-quality regions of the search space, which is needed to identify the optimal ST-block. 

Specifically, we randomly sample $c$ $combs$ from the remaining search space in the $i$-th pruning stage.
For each $comb$, we randomly sample $r$ ST-blocks and train the $j$-th ST-block on the $j$-th task, resulting in the training set $S_i = \displaystyle\bigcup_{k=1}^c\{(t_j,b_j,R(t_j,b_j))\}_{j=1,\cdots,r}^{b_j \in comb_k}$, which is used to calculate EDF values for the sampled $combs$ using Equation~\ref{eq: EDF} and is also used to pretrain the TAP.
After the pruning is completed, we randomly sample $z$ ST-blocks in the final pruned search space and train them on the $r$ ($r<z$) tasks to construct a training set $S_e = \{(t_j,b_k,R(t_j,b_k)\}_{j=1,\cdots,r} ^{ k=1,\cdots,z}$.
%
Finally, we pretrain the TAP on the training set $\displaystyle\bigcup_{i=1}^p S_i \cup S_e$, where $p$ is the number of pruning stages.


\subsubsection{Zero-shot Search on Arbitrary Unseen CTS Forecasting Tasks}
\label{sssec: zero}

After the pretraining, we obtain a TAP that can perform zero-shot search on arbitrary unseen CTS forecasting tasks.
Given an unseen task $t^\prime$, we first perform the iterative pruning to reduce the general search space to a high-quality search space customized for task $t^\prime$ and then use the pretrained TAP to traverse the final pruned search space to find the optimal ST-block.

Specifically, in the $i$-th pruning stage, we first randomly sample $c$ combs from the remaining search space. For each $comb$, we randomly sample $r$ ST-blocks and train them on task $t^\prime$, resulting in the training set $S_i^\prime = \displaystyle\bigcup_{k=1}^c\{(t^{\prime},b_j,R^\prime(t_j,b_j))\}_{j=1,\cdots,r}^{b_j \in comb_k}$, where $R^\prime(t, b)$ is the prediction accuracy ($\widehat{MAE}$) of ST-block $b$ on task $t$ generated by the pretrained TAP.
We then use $S_i^\prime$ to calculate pseudo-EDF ($\widehat{EDF}$) labels for the sampled $combs$ using Equation~\ref{eq: EDF} and train a $comb$ predictor to prune the search space.

After the pruning is completed, we obtain a high-quality search space customized for task $t^\prime$.
We then use the pretrained TAP to predict the accuracy of all ST-blocks in the pruned search space and rank them according to prediction accuracy, where the top-1 ST-block is returned as the optimal ST-block.
Since the zero-shot search phase does not involve any training, it is very fast and can be completed in minutes.



\subsection{Fast Training with Parameter Adaptation}
\label{ssec: adapt}
%
Having found {a high-performance} ST-block, we train its parameter weights to make forecasts on unseen tasks.
Unlike existing automated methods that regard the training of different ST-blocks as independent processes and train the identified ST-block from scratch, we inherit the parameter weights of pretrained ST-blocks to \textbf{accelerate the training} of the identified ST-block.

\subsubsection{Motivation for Parameter Weights Inheritance}
In the pretraining phase, we train a large number of ST-blocks to generate training samples for TAP. The parameter weights of these ST-blocks contain knowledge about how to capture general spatio-temporal patterns of CTS, which is shared across ST-blocks and tasks.
Inheriting this knowledge when training identified ST-blocks may facilitate fast convergence because the identified ST-blocks are trained from a good starting point rather than from scratch.

We inherit knowledge in pretrained ST-blocks by inheriting their parameter weights, unlike common transfer learning methods. Since pretrained ST-blocks differ in architecture from the identified ST-block and are trained on different tasks, it is not feasible to copy their parameter weights directly to the identified ST-block. However, the parameters of an ST-block come from multiple S/T operators that make up the ST-block, and the number of pretrained ST-blocks is sufficiently large to include all S/T operators in the search space, so we can inherit the parameter weights of each S/T operator in the identified ST-block separately.

\begin{figure}[b]
  \centering
  \includegraphics[width=0.78\linewidth]{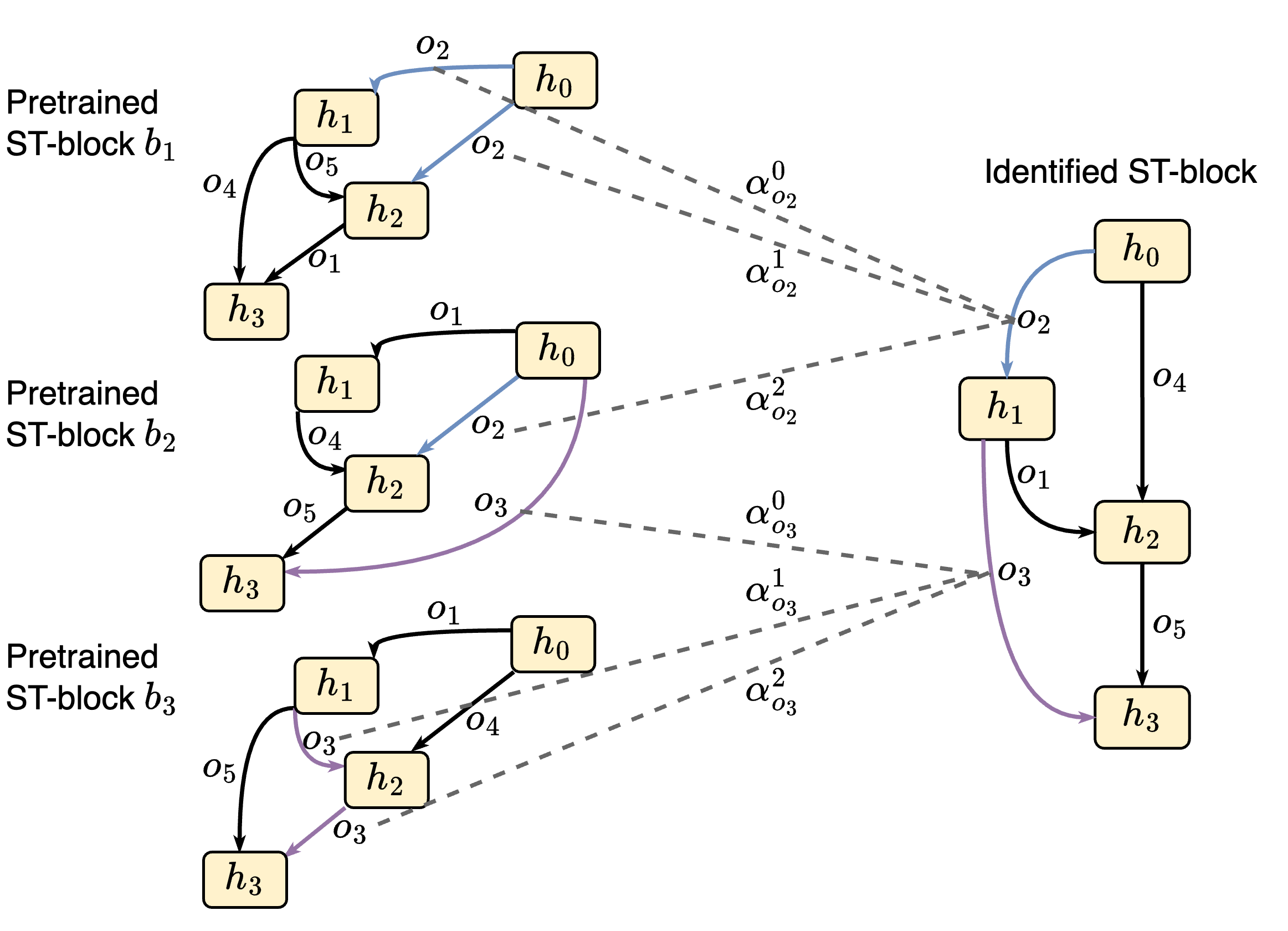}
  \caption{Inheriting parameters from pretrained ST-blocks.
  }
  \label{fig: weight}
\end{figure}

%
\subsubsection{Fast Parameter Adaptation}
We propose a fast parameter adaptation strategy to inherit the weights of pretrained ST-blocks in a learnable manner. Then we finetune the inherited weights on the target task for a few training steps to obtain optimal weights for the identified ST-block.

Specifically, for each S/T-operator $o$ in the identified ST-block, we first traverse the pretrained ST-blocks and select $n$ ST-blocks that contain $o$ and are most similar to the identified ST-block.
Here, we consider two approaches to calculating the similarity between two ST-blocks. One is graph edit distance between DAGs corresponding to the two ST-blocks, which captures structural similarity. The other is the cosine similarity between the architectural feature vectors output by AFL (see Figure~\ref{fig: TAP}), which captures semantic similarity.
We combine the two through voting.

Then, for each S/T operator $o$, we introduce $k$ learnable vectors $\alpha_o=\{\alpha_o^1, \alpha_o^2, ..., \alpha_o^k\}$, where $\alpha_o^i \in \mathbb{R}^{n_o}$ and $n_o$ is the number of parameters in $o$, to linearly transform the parameter weights of $o$ that from $n$ selected pretrained ST-blocks. Then we calculate the average of the transformed weights as the initial parameter weights of the identified ST-block, formulated as follows.
\begin{align}
W_{o}^{init} = (\alpha_o^1 W_{o}^{1} + \alpha_o^2 W_{o}^{2} + ... + \alpha_o^k W_{o}^{k}) / k,
\end{align}
where $W_{o}^{i}$ is the $i$-th parameter weight vector of $o$, $k$ is the number of $o$ contained in $n$ selected pretrained ST-blocks, and $W_{o}^{\mathit{init}}$ is the initial parameter weights of $o$ of the identified ST-block.
Figure~\ref{fig: weight} illustrates the proposed fast parameter adaptation strategy, where operator $o_2$ in the identified ST-block inherits parameter weights $W_{o_2}^{1}$, $W_{o_2}^{2}$, and $W_{o_2}^{3}$ from ST-blocks $b_1$ and $b_2$, and the corresponding coefficient vectors are $\alpha_{o_2}^1$, $\alpha_{o_2}^2$ and $\alpha_{o_2}^3$, respectively.
{We further divide $W_o^i$ into groups based on the neural modules they belong to, with the coefficients being shared within each group to reduce the number of learnable parameters in $\alpha_o^i$.
%
}

The coefficient $\alpha$ for all operators in the identified ST-block are learned through the following objective, enabling adaptive inheritance of knowledge from pretrained ST-blocks.
\begin{align}
\mathit{argmin}_{\alpha} \mathbb{E}_{x\sim D} \mathcal{L}(x; W^{\mathit{init}}(\alpha, W)),
\label{eq: alpha}
\end{align}
where $W$ contains the parameter weights of the pretrained ST-blocks, $D$ is the target task, and $\mathcal{L}$ is the MAE loss metric.
The idea behind this optimization objective is to learn a set of coefficients to linearly transform the pretrained weights to instantiate the weights of the identified ST-block such that it achieves the highest performance on the target dataset.

%

%


%
After we obtain $W^{\mathit{init}}$, we further finetune it on the target dataset to get the final weights $W_{\mathit{final}}$ of the identified ST-block as,
\begin{align}
W^{\mathit{final}} = \mathit{argmin}_{W^{init}} \mathbb{E}_{x\sim D} \mathcal{L}(x; W^{init})
\end{align}
Since we train the identified ST-block from a good starting point $W^{init}$, it converges much faster than when training it from scratch, thus achieving the goal of reducing the training time.


\section{Experiments}
We conduct comprehensive experiments on seven public CTS forecasting datasets to assess the effectiveness and efficiency of FACTS.

\subsection{Experimental Setup}
\label{ssec: expsetup}

\subsubsection{Tasks for Pretraining and Evaluating}
We summarize the datasets used for pretraining and evaluation below.

\noindent
\textbf{Datasets and tasks for pretraining: } METR-LA~\cite{li2018dcrnn_traffic}, ETTh1~\cite{haoyietal-informer-2021}, ETTh2~\cite{haoyietal-informer-2021}, ETTm1~\cite{haoyietal-informer-2021}, ETTm2~\cite{haoyietal-informer-2021}, Solar-Energy~\cite{lai2018modeling}, ExchangeRate~\cite{lai2018modeling}, PEMS03~\cite{song2020spatial}, PEMS04~\cite{song2020spatial}, PEMS07~\cite{song2020spatial}, PEMS08~\cite{song2020spatial}.
We form 200 CTS forecasting tasks based on these datasets by splitting them along the temporal or spatial dimensions and considering the forecasting settings $P$-12/$Q$-12 and $P$-48/$Q$-48.

\noindent
\textbf{Datasets and tasks for evaluation: } Electricity~\cite{lai2018modeling}, NYC-TAXI~\cite{ye2021coupled}, NYC-BIKE~\cite{ye2021coupled}, SZ-TAXI~\cite{zhao2019t}, Los-Loop~\cite{zhao2019t}, PEMS-BAY~\cite{li2018dcrnn_traffic}, PEMSD7(M)~\cite{yu2018spatio}.
%
We create 28 unseen tasks by considering the forecasting settings $P$-12/$Q$-12, $P$-24/$Q$-24, $P$-48/$Q$-48, and $P$-168/$Q$-1 (3rd) for each of these datasets.



\subsubsection{Baselines}
We compare FACTS with three competitive manually designed and three automated baselines. We reproduce the baselines based on their released code.

\begin{itemize}[leftmargin=*]
    \item[$\bullet$] MTGNN: employs mix-hop graph convolution and dilated inception convolution to build ST-blocks~\cite{wu2020connecting}.
    \item[$\bullet$] AGCRN: employs 1D GCNs and GRUs to build ST-blocks~\cite{bai2020adaptive}.
    \item[$\bullet$] PDFormer: employs spatial and temporal transformers to build ST-blocks~\cite{jiang2023pdformer}.
    \item[$\bullet$] AutoCTS: A gradient-based automated CTS forecasting framework with a manually designed search space~\cite{wu2022autocts}.
    \item[$\bullet$] AutoCTS+: A comparator-based automated CTS foreacasting framework that jointly searches for ST-blocks and accompanying hyperparameter settings~\cite{wu2023autocts+}.
    \item[$\bullet$] SimpleSTG: prunes the general search space manually using sampling to remove bad design choices and employs random search to find the optimal ST-block~\cite{xu2022understanding}.
\end{itemize}

\subsubsection{Evaluation Metrics}
Following previous studies~\cite{yu2018spatio,li2018dcrnn_traffic,DBLP:conf/ijcai/WuPLJZ19,bai2020adaptive,wu2020connecting}, we use mean absolute error (MAE), root mean squared error (RMSE), and mean absolute percentage error (MAPE) as the evaluation metrics for multi-step forecasting, and we use Root Relative Squared Error (RRSE) and Empirical Correlation Coefficient (CORR) as evaluation metrics for single-step forecasting. For MAE, RMSE, MAPE, and RRSE, lower values are better, while for CORR, higher values 
are better.

\subsubsection{Implementation Details}
\noindent 
\\\textbf{$comb$ predictor.}
The $comb$ predictor is a GBDT model based on LightGBM~\cite{ke2017lightgbm} with 100 trees and 31 leaves per tree. Standard normalization is applied to normalize the EDF labels of $combs$. We train the $comb$ predictor with a learning rate of 0.05 and MSE loss.

\noindent \textbf{TAP.}
For the structure of TAP, we set the number of layers of the GCNs to 4, with 128 hidden units in each layer.
To pretrain TAP, we use the Adam~\cite{kingma2014adam} optimizer with a learning rate of 0.001 and a weight decay of 0.0005. The batch size is set to 64. Moreover, the hidden dimensions of the two-layer set-transformer are set to 256 and 128, respectively. We extract 128 commonly-used statistical features. We apply min-max normalization to normalize the validation MAE of ST-blocks from the same task and train TAP for 100 epochs with an early stop patience of 5 epochs.

\noindent \textbf{Pruning.}
We randomly sample $c=100$ $combs$ at each pruning stage and collect $r=100$ ST-blocks for each $comb$. The pruning ends when the number of $combs$ in the remaining search space is less than $M=2,000$. We then collect $z=10,000$ ST-blocks in the pruned search space. 

\noindent\textbf{Parameter adaptation.}
We select $n=5$ ST-blocks most similar to the identified ST-block for parameter adaptation.

\noindent \textbf{CTS forecasting models.}
We use MAE as the training objective to train CTS forecasting models, and use Adam with a learning rate of 0.001 and a weight decay of 0.0001 as the optimizer. We set the batch size to 64.  
For fair comparison, we train all CTS forecasting models for 100 epochs.




\begin{table*}[!htbp]
\small
    \centering
    \caption{Performance of P-12/Q-12 and P-24/Q-24 forecasting.} %
    \resizebox{0.95\textwidth}{!}{
        \begin{tabular}{c|c|ccccccc|ccccccc}
        \hline
        &&\multicolumn{7}{c|}{P-12/Q-12 forecasting}&\multicolumn{7}{c}{P-24/Q-24 forecasting} \\
        Data & Metric & FACTS & \emph{AutoCTS} & \emph{AutoCTS+}&\emph{SimpleSTG} & MTGNN & AGCRN & PDFormer& FACTS & \emph{AutoCTS} & \emph{AutoCTS+}&\emph{SimpleSTG} & MTGNN & AGCRN & PDFormer \\
        \hline
        \multirow{3}{*}{PEMS-BAY} & MAE &\textbf{1.517}  & 1.736 & \underline{1.572} &1.815 & 1.960 & 1.652  & 1.742 &\textbf{1.780}  & 1.911 & \underline{1.836} &1.877 & 1.884 & 2.124 & 1.843\\
        & RMSE &\textbf{3.287}  & 3.935 & \underline{3.531} &4.281  & 4.461 & 3.815 & 3.920 &\textbf{3.986} & 4.435 & \underline{4.017} &4.291 & 4.331 & 4.612 & 4.112 \\
        & MAPE &\textbf{3.494\%}  & 3.822\% & \underline{3.501\%} &4.316\% & 4.560\% & 3.843\% & 3.947\% &\textbf{4.102\%} &4.784\% & 4.236\% &4.298\% & 4.405\% & 5.113\% & \underline{4.139\%}\\
        \hline
        \multirow{3}{*}{Electricity} & MAE &\textbf{225.814} & 240.65 & \underline{238.513} &262.815  & 306.331 & 611.08 & 247.982 &\textbf{193.786} & 204.333 & 205.401 &207.185 & 211.913 & 1718.216 & \underline{202.571}\\
        & RMSE &\textbf{1943.750}  & 2177.910 & \underline{2106.307} &2329.154 & 2468.959 & 8288.991 & 2178.527 &\textbf{1676.211} &\underline{1681.030} & 1784.313 &1825.164  & 1871.110 & 16364.798 & 1724.646 \\
        & MAPE &\textbf{15.871\%} & 17.275\% & 16.961\% &21.165\%  & 24.381\% & 42.628\% & \underline{16.864\%} &\textbf{15.468\%} &\underline{15.790\%} &16.085\% &16.581\%  &17.136\% &58.626\% &15.993\%\\
        \hline
        \multirow{3}{*}{PEMSD7M} & MAE &\textbf{2.551}  & 2.604 & 2.617 &\underline{2.592} & 2.643 & 2.697 & 2.631 &\textbf{3.167} &3.227 & \underline{3.191} &3.245 & 3.327 & 8.823 & 3.310\\
        & RMSE &\textbf{4.863}  & 5.195 & 5.166 &\underline{5.018}  & 5.217 & 5.401 & 5.261 &\textbf{6.078} &\underline{6.171} &6.221 &6.316 & 6.315 & 14.674 & 6.208\\
        & MAPE &\textbf{6.296\%}  & 6.592\% & 6.581\% &\underline{6.471\%}  & 6.523\% & 6.782\% & 6.482\% &\textbf{8.112\%} &8.328\% &8.323\% &8.359\%  &8.493\% &28.915\% &\underline{8.251\%}\\
        \hline
        \multirow{3}{*}{NYC-TAXI} & MAE &\textbf{5.334}  & 5.576 & \underline{5.536} &5.625  & 5.847 & 5.818 & 6.259 &\textbf{5.550} & 5.621 & \underline{5.591} &5.774  & 5.889 & 5.735 & 5.760\\
        & RMSE &\textbf{9.472 } & 9.846 & \underline{9.792} &10.048  & 11.918 & 13.924 & 11.206 &\textbf{9.995} &10.834 & \underline{10.127} &10.830 &10.710 &11.512 &10.560\\
        & MAPE &\textbf{37.780\%}  & \underline{39.985\%} & 41.235\% &42.011\%  & 40.271\% & 43.027\% & 43.194\% &\textbf{38.338\%}  &\underline{38.452\%} &40.174\% &42.915\%  &39.914\% &44.264\% &43.709\%\\
        \hline
        \multirow{3}{*}{NYC-BIKE} & MAE &\textbf{1.796}  & 1.891 & \underline{1.811}&1.832 & 1.942 & 1.856 & 2.368 &\textbf{1.842} &\underline{1.887} &1.988 &1.895  &2.241 &1.976 & 2.049\\
        & RMSE &\textbf{2.732}  &2.992  &\underline{2.794} &2.815  & 3.265 & 3.019 & 3.625 &\textbf{2.867} &3.140 &3.232 &3.274  &3.310 &\underline{3.101} & 3.147 \\
        & MAPE &\textbf{49.951\%} & 52.389\% & \underline{52.214\%} &52.281\%  & 53.322\% & 56.128\% & 54.161\% &\textbf{50.675\%} &50.915\% &51.313\% &\underline{50.781\%}  &54.015\% &57.315\% &52.648\%\\
        \hline
        \multirow{3}{*}{Los-Loop} & MAE &\textbf{3.578}  & 3.677 & 3.652 &3.668  & \underline{3.640} & 8.912 & 3.980 &\textbf{4.101} &\underline{4.179} &4.244 &4.375  &4.271 &4.452 &4.282\\
        & RMSE &\textbf{6.841} & 7.063 & 7.119 &7.179  & \underline{7.084} & 15.182 & 7.286 &\textbf{7.518} &\underline{7.775} &7.943 &8.143  &7.905 &8.831 &8.014\\
        & MAPE &\textbf{10.006\%} & 10.720\% & 10.443\% &10.609\% & \underline{10.212\%} & 34.040\% & 10.668\% &\textbf{12.279\%} &\underline{12.857\%} &13.209\% &13.591\%  &12.876\% &13.720\% &12.945\% \\
        \hline
        \multirow{2}{*}{SZ-TAXI} & MAE &\textbf{3.178} & 3.250 & 3.254 &\underline{3.218} &3.229 & 4.510 & 3.719 &\textbf{2.892} &3.171 &3.056 &3.154 &3.215 &\underline{2.905} &3.237\\
        & RMSE &\textbf{4.217}  & 4.484 & 4.457 &\underline{4.291} & 4.779 & 5.002 & 4.904 &\textbf{4.260} &4.459 &\underline{4.347} &4.401 &4.528 &4.401 &4.542 \\
        \hline
      \end{tabular}
    }
    \label{tab: 12-12}
\end{table*}

\begin{table*}[!htbp]
\small
    \centering
    \caption{Performance of P-48/Q-48 and P-168/Q-1(3rd) forecasting.} %
    \resizebox{0.95\textwidth}{!}{
        \begin{tabular}{c|c|ccccccc|c|ccccccc}
        \hline
        &&\multicolumn{7}{c|}{P-48/Q-48 forecasting}&&\multicolumn{7}{c}{P-168/Q-1(3rd) forecasting} \\
        Data & Metric & FACTS & \emph{AutoCTS} & \emph{AutoCTS+}&\emph{SimpleSTG} & MTGNN & AGCRN & PDFormer& Metric & FACTS & \emph{AutoCTS} & \emph{AutoCTS+}&\emph{SimpleSTG} & MTGNN & AGCRN & PDFormer \\
        \hline
        \multirow{3}{*}{PEMS-BAY} & MAE &\textbf{1.963}  &1.988  &\underline{1.974} &2.011 & 2.198 & 2.831 & 2.064 & RRSE &\textbf{0.2887}  &\underline{0.2901} &0.2931 &0.3015  &0.2947 &0.4719 &0.2940\\
        & RMSE &\textbf{4.175}  &\underline{4.185}  &4.253 &4.291 &4.684 & 5.125 & 4.392 & CORR &\textbf{0.9366} &\underline{0.9275} &0.9241 &0.8906  &0.9116 &0.8586 &0.9245\\
        & MAPE &\textbf{4.708\%}  & \underline{4.715\%} & 4.856\% &4.913\% & 5.330\% & 5.013\% &4.928\% & \\
        \hline
        \multirow{3}{*}{Electricity} & MAE &\textbf{215.821}  & 247.256 & \underline{239.836} &254.150 & 306.331 & 611.08 & 250.982 &RRSE &\textbf{0.0692} &0.0741 &\underline{0.0731} &0.0815 &0.0758 &0.1033 &0.0781\\
        & RMSE &\textbf{1839.745} & 2144.362 & \underline{2038.137} &2249.158  & 2468.959 & 8288.991 & 2218.027 &CORR &\textbf{0.9785} &0.9439 &\underline{0.9516} &0.8911 &0.9412 &0.8854 &0.9273\\
        & MAPE &\textbf{16.438\%}  & 16.845\% & \underline{16.761\%} &16.853\% & 24.381\% & 42.628\% & 16.864\%  &\\
        \hline
        \multirow{3}{*}{PEMSD7M} & MAE &\textbf{3.454} & 3.522 & 3.510&\underline{3.471}  &3.585 &3.606 &3.559 & RRSE &\textbf{0.2867} &0.3056 &\underline{0.2895} &0.2981  &0.3105 &0.5314 &0.2995 \\
        & RMSE &\textbf{6.582}  & 6.715 & 6.671 &\underline{6.621} & 6.907 & 7.124 & 6.814 &CORR &\textbf{0.9375} &0.9298 & \underline{0.9338} &0.9282  & 0.9278 &0.8186 &0.9336\\
        & MAPE &\textbf{8.767\%}  &9.236\% & 9.114\% &\underline{8.991\%} & 9.322\% & 9.375\% & 9.201\% & \\
        \hline
        \multirow{3}{*}{NYC-TAXI} & MAE &\textbf{5.544}  & 5.956 & \underline{5.622} &5.811  & 5.767 & 6.009 &5.995 & RRSE & \textbf{0.2018} &0.2324 &\underline{0.2191} &0.2571 &0.2273 &0.2709 &0.2624 \\
        & RMSE &\textbf{9.931}  & 11.154 & \underline{10.174} &10.781 & 10.568 & 18.049 & 11.963 & CORR &\textbf{0.8874} &0.8689 &\underline{0.8714} &0.8491  &0.8697 &0.8473 &0.8417\\
        & MAPE &\textbf{37.964\% } & 39.156\% & \underline{38.235\%} &38.681\%  & 39.011\% & 49.629\% & 39.894\% & \\
        \hline
        \multirow{3}{*}{NYC-BIKE} & MAE &\textbf{1.990} &\underline{1.998} & 2.026 &2.184  & 2.097 & 2.105 & 2.096 & RRSE &\textbf{0.7478} &\underline{0.7492} &0.7515 &0.7590 &0.7581 &0.7601 &0.7531\\
        & RMSE &\textbf{2.996}  &\underline{3.011} & 3.178 &3.527 & 3.256 & 3.311 & 3.308 &CORR &\textbf{0.7850} &\underline{0.7805} &0.7773 &0.7685  & 0.7691 &0.7631 & 0.7669\\
        & MAPE &\textbf{51.830\% } & \underline{52.146\%} & 52.471\% &55.107\% & 52.737\% & 53.481\% & 52.672\% &\\
        \hline
        \multirow{3}{*}{Los-Loop} & MAE &\textbf{4.520} &4.643 &\underline{4.551} &4.750 &4.624 &8.962& 4.767 & RRSE &\textbf{0.4258} &0.4311 &\underline{0.4298}&0.4307 &0.4392 &1.7565 &0.4493\\
        & RMSE &\textbf{8.327} &8.785 &\underline{8.536} &8.681  &8.549 &14.956 &9.003 & CORR &\textbf{0.7908} &0.7797 &\underline{0.7825} &0.7783  &0.7695 &0.0203 &0.7699\\
        & MAPE &\textbf{14.873\%} &15.394\% & \underline{15.015\%} &15.491\%  &17.417\% &34.184\% &15.961\% &\\
        \hline
        \multirow{2}{*}{SZ-TAXI} & MAE &\textbf{2.891} &3.106 &\underline{2.976}&3.015 & 3.193 &2.980 &3.196 & RRSE &\textbf{0.2288} &0.4815 &\underline{0.4785}&0.4916 &0.4878 &1.2367 &0.4892  \\
        & RMSE &\textbf{4.394} &4.471 &4.403 &4.455 &4.486 &\underline{4.397} &4.496 &CORR &\textbf{0.5077} &0.2206 &0.2200 &0.2003 &0.2216 &0.0345 &\underline{0.2236} \\
        \hline
      \end{tabular}
    }
    \label{tab: 168-3}
\end{table*}

\subsection{Experimental Results}

\subsubsection{Main Results}

Tables~\ref{tab: 12-12} and~\ref{tab: 168-3} present the overall performance of FACTS and the baselines on the seven unseen CTS forecasting datasets with different forecasting settings.
We run all evaluations three times with different random seeds and report the mean numbers.
For ease of observation, we use bold and underline to highlight the best and second-best results, respectively.
We observe that: 1) FACTS consistently outperforms the existing manual and automated methods, although we never pretrain on the $P$-24/$Q$-24 and $P$-168/$Q$-1 (3rd) forecasting settings and the seven datasets. This is evidence of the effectiveness of FACTS on arbitrary unseen CTS forecasting tasks.
2) AutoCTS+ usually achieves the second best accuracy because it supports joint search of ST-block and accompanying hyperparameter values. 
However, FACTS still outperforms AutoCTS+, although it does not search for hyperparameter values.
Additionally, joint search can easily be integrated into FACTS, which may further improve performance.

\subsubsection{Effectiveness of Automated Search Space Pruning}

We design comparison experiments to assess the effectiveness of the automated search space pruning.

\noindent
\\\textbf{Pruning improves the quality of the search space.}
First, we find that our pruning strategy can improve the quality of the search space.
We consider 6 variants of FACTS that perform only 0 to 5 pruning stages, resulting in 6 pruned search spaces: space-0, space-1, space-2, space-3, space-4, and space-5. FACTS performs 6 pruning stages.
We also consider the manually designed search spaces of AutoCTS and SimpleSTG.
We randomly sample 200 ST-blocks from each of the above search spaces and train them to obtain their validation accuracy and EDF values.

\begin{figure}[!htbp]
  \centering
  \includegraphics[width=0.95\linewidth, trim=80 60 60 50,clip]{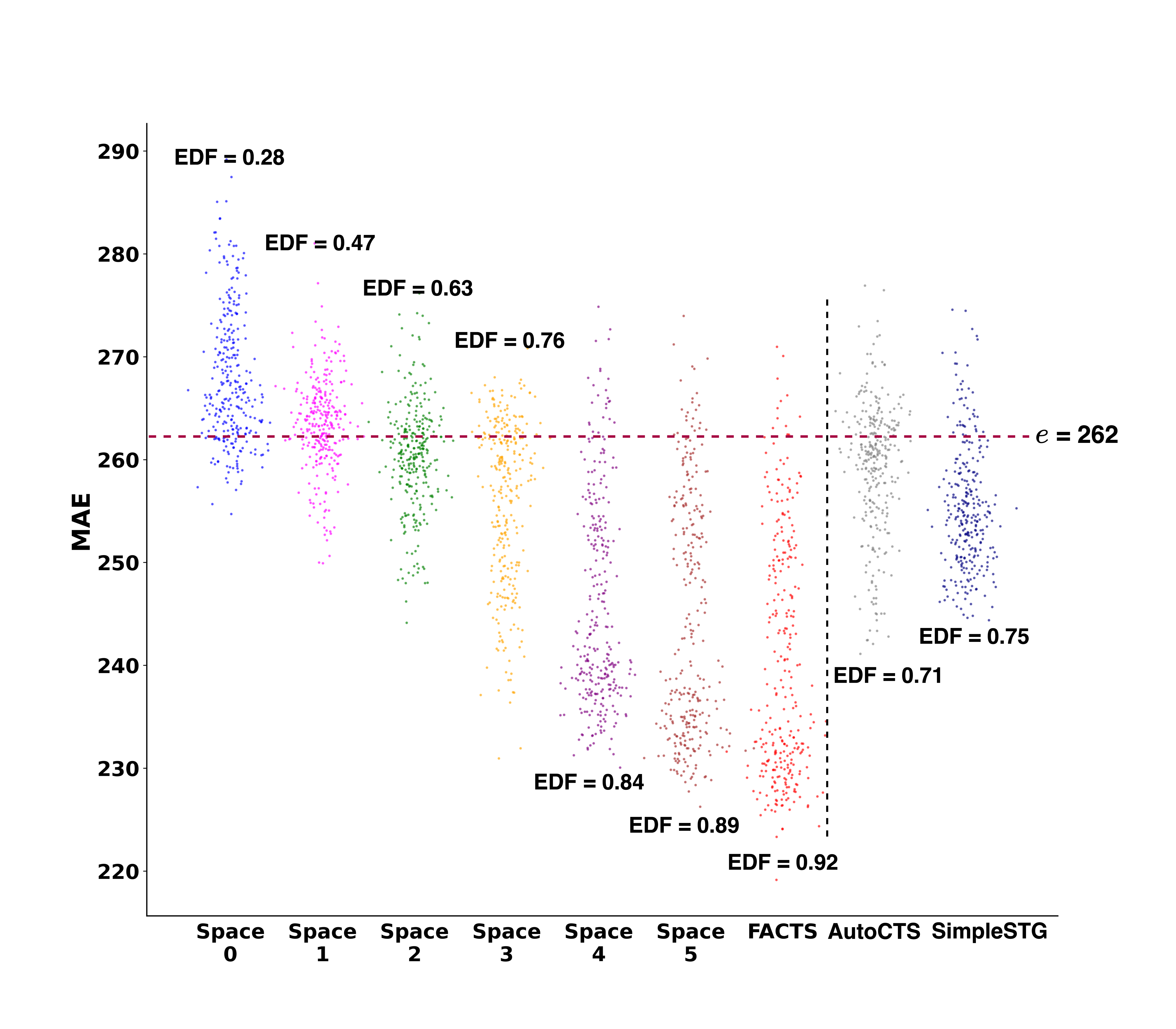}
  \caption{Quality of the search spaces during pruning.
  }
  \label{fig: pruning}
\end{figure}

%
We draw a scatter plot to show the results on Electricity under the $P$-12/$Q$-12 forecasting setting.
Figure~\ref{fig: pruning} shows that with the pruning of the search space, the sampled ST-blocks show increasingly higher accuracy, which is evidence that the proposed search space pruning strategy can remove low-quality regions of the search space, thereby improving the quality of the remaining search space.
Further, the quality of the search spaces of AutoCTS and SimpleSTG are higher than that of FACTS in the early stages, but are surpassed after 4 iterations of the pruning, which indicates that the automated pruning strategy can generate search spaces that are better than the manually designed search spaces on unseen tasks.

\noindent
\textbf{Pruning helps search for high-performance ST-blocks.} Second, we demonstrate that our pruning strategy helps search for high-performance ST-blocks.
We compare FACTS with the variants space-0 to space-5.
For fair comparison, we randomly sample $m$ ST-blocks in each search space above, where $m$ is the size of the final pruned search space of FACTS, which in this case is equal to $28,206,080$, and we predict them using the pretrained TAP to find the optimal ST-block, and then we train it to obtain the accuracy.

We show the results on the Electricity dataset under the $P$-12/$Q$-12 forecasting setting in Table~\ref{tab: ablation}. We see that the identified ST-blocks show increasingly higher accuracy as the search space is pruned, which indicates that the iterative pruning strategy gradually improves the quality of the pruned search space, making it easier to find high-performance ST-blocks.

\noindent
\textbf{Iterative pruning v.s. one-time pruning.}
We offer evidence that iteratively pruning the search space yields better performance than pruning the search space only once (see Section~\ref{iter}).
We design three variants $one$-$time$-$1$, $one$-$time$-$2$, $one$-$time$-$3$, which prune the search space only once, and sample the same number of ($comb$, EDF) pairs as FACTS to train a $comb$ predictor, and prune the search space to $M$ combs in one step.
The EDF thresholds of the three variants are set to the initial, intermediate and final values of the EDF thresholds of FACTS, respectively.

Table~\ref{tab: w/o iterative pruning} shows that although we set different EDF thresholds for the three variants, they all perform worse than FACTS.
This shows the effectiveness of the proposed iterative pruning strategy.

\begin{table}[!htbp]
    \small
    \centering
    \caption{Accuracy comparison in the pruning process.}
    \resizebox{0.9\linewidth}{!}{
    \begin{tabular}{c|cccccccc}
        \hline
        \multirow{1}{*}{Metric} & space-0 & space-1 & space-2 & space-3 & space-4 & space-5 & FACTS\\
        \hline
        MAE & 258.540 & 250.515 & 242.858 &  235.870 & 231.231 &227.410 & \textbf{225.814}\\
        RMSE & 2077.255 & 2050.231 & 2016.333 & 1990.970 & 1976.011 &1955.121 & \textbf{1943.750}\\
        MAPE & 24.918\% & 22.442\% & 19.892\% & 17.932\% & 17.004\% &16.115\% &\textbf{15.871\%}\\
        \hline
    \end{tabular}
    }
    \label{tab: ablation}
\end{table}

\begin{table}[t]
    \caption{Iterative pruning v.s. one-time pruning.}
    \label{tab: w/o iterative pruning}
     \centering
     \resizebox{0.60\linewidth}{!}{

    \begin{tabular}{cc|ccc}
            \hline
             variant &metric & PEMS-BAY & NYC-TAXI & Electricity \\ \hline
           \multirow{3}{*}{$\textit{One-time-1}$} & MAE &1.637  &5.615  &256.814 \\
              & RMSE &3.725  &10.347  &2279.015  \\
             & MAPE  &3.712\%  &41.925\%  &18.517\%  \\
           \multirow{3}{*}{$\textit{One-time-2}$} & MAE &1.574  &5.581  &245.618\\
              & RMSE &3.549  &9.991  &2185.761  \\
             & MAPE  &3.682\%  &41.034\%  &16.880\%  \\
            \multirow{3}{*}{$\textit{One-time-3}$} & MAE &1.762  &5.704  &273.047\\
              & RMSE &4.113  &10.675  &2365.225  \\
             & MAPE  &4.042\%  &43.714\%  &23.027\%  \\
             \multirow{3}{*}{FACTS} & MAE &\textbf{1.517}  &\textbf{5.334}  &\textbf{225.814}  \\
             & RMSE &\textbf{3.287} &\textbf{9.472} &\textbf{1943.750}\\
             & MAPE &\textbf{3.494\%} &\textbf{37.780\%} &\textbf{15.871\%} \\
             \hline
             
        \end{tabular}
        }
\end{table}

\noindent

\textbf{{Performance comparison between EDF and mean MAE.}}
{We compare FACTS and the variant that replaces EDF with the mean MAE metric.}

{The results in Table~\ref{tab: mae res-24} show that FACTS performs considerably better than the variant on all CTS forecasting tasks, indicating that EDF is the better metric when evaluating the quality of a $comb$.}

\begin{table}[!htbp]
\small
    \centering
    \caption{{Performance comparison of EDF vs.\ mean MAE.}} %
    \resizebox{\linewidth}{!}{{
        \begin{tabular}{c|c|cc|cc|cc|c|cc}
        \hline
        &&\multicolumn{2}{c|}{P-12/Q-12} &\multicolumn{2}{c|}{P-24/Q-24} &\multicolumn{2}{c|}{P-48/Q-48} & &\multicolumn{2}{c}{P-168/Q-1(3rd)}\\
        Data & Metric & EDF & MAE& EDF & MAE&EDF & MAE&Metric& EDF & MAE \\
        \hline
        \multirow{3}{*}{PEMS-BAY} & MAE & \textbf{1.517} & 1.572&\textbf{1.780} &1.869   &\textbf{1.963} & 1.976 &RRSE& \textbf{0.2887} & 0.2933\\
        & RMSE & \textbf{3.287} & 3.557&\textbf{3.986}&4.115  &\textbf{4.175} & 4.192 &CORR & \textbf{0.9366} & 0.9244 \\
        & MAPE &\textbf{3.494\%} & 3.515\%&\textbf{4.102\%}&4.148\%  &\textbf{4.708\%} & 4.732\%  & & &\\
        \hline
        \multirow{3}{*}{Electricity} & MAE & \textbf{225.814} & 265.140 &\textbf{193.786} &208.517  &\textbf{215.821} & 258.191 &RRSE& \textbf{0.0692} & 0.0744\\
        & RMSE &\textbf{1943.750} & 2237.214&\textbf{1676.211}&1715.150  &\textbf{1839.745}  & 2025.159 &CORR& \textbf{0.9785} & 0.9418 \\
        & MAPE &\textbf{15.871\%} &16.965\% &\textbf{15.468\%}&16.005\%  &\textbf{16.438\%}  &17.002\% & & & \\
        \hline
        \multirow{3}{*}{PEMSD7M} & MAE &\textbf{2.551} & 2.602 &\textbf{3.167} &3.341  &\textbf{3.454} & 3.498 & RRSE & \textbf{0.2867} & 0.2893\\
        & RMSE &\textbf{4.863} &5.025 &\textbf{6.078}&6.371  &\textbf{6.582} & 6.628 &CORR&\textbf{0.9375} & 0.9326\\
        & MAPE &\textbf{6.296\%} &6.482\%&\textbf{8.112\%}&8.419\%  &\textbf{8.767\% } &8.994\% & & &\\
        \hline
        \multirow{3}{*}{NYC-TAXI} & MAE & \textbf{5.334} & 5.558&\textbf{5.550} &5.701   &\textbf{5.544}  & 5.629 &RRSE & \textbf{0.2018} & 0.2292\\
        & RMSE &\textbf{9.472} &9.801&\textbf{9.995}&10.332   &\textbf{9.931} &10.255 & CORR &\textbf{0.8874} &0.8709\\
        & MAPE &\textbf{37.780\%}&40.112\% &\textbf{38.338\%}&40.156\%   &\textbf{37.964\%} &38.414\%  & & &\\
        \hline
        \multirow{3}{*}{NYC-BIKE} & MAE &\textbf{1.796} &1.821 &\textbf{1.842}&1.908   &\textbf{1.990}  &2.004 &RRSE &\textbf{0.7478} & 0.7554\\
        & RMSE &\textbf{2.732} &2.835 &\textbf{2.867}&3.185  &\textbf{2.996}  &3.143 &CORR &\textbf{0.7850} & 0.7785 \\
        & MAPE &\textbf{49.951\%} &52.281\% &\textbf{50.675\%}&53.161\%  &\textbf{51.830\%}  &52.315\% & & &\\
        \hline
        \multirow{3}{*}{Los-Loop} & MAE &\textbf{3.578} &3.637 &\textbf{4.101}&4.298   &\textbf{4.520}  &4.582 &RRSE &\textbf{0.4258} &0.4302\\
        & RMSE &\textbf{6.841} &7.091 &\textbf{7.518}&7.801  &\textbf{8.327}  &8.522 &CORR &\textbf{0.7908} &0.7798\\
        & MAPE &\textbf{10.006\%} &10.285\% &\textbf{12.279\%}&13.055\%  &\textbf{14.873\%}  &15.296\% & & & \\
        \hline
        \multirow{2}{*}{SZ-TAXI}& MAE &\textbf{3.178} &3.229 &\textbf{2.892}&3.161 &\textbf{2.891} &3.115 &RRSE &\textbf{0.2288} &0.4382\\
        & RMSE &\textbf{4.217} &4.320 &\textbf{4.260}&4.398  &\textbf{4.394} &4.477&CORR &\textbf{0.5077} &0.2815 \\
        \hline
      \end{tabular}
    }
    }
    \label{tab: mae res-24}
\end{table}

%

\subsubsection{Effectiveness of Zero-shot Search Strategy}

\noindent
\\\textbf{Search time comparison.} We compare the search time of FACTS and the automated baselines on seven unseen datasets under the $P$-24/$Q$-24 forecasting setting and report the results in Table~\ref{tab: time1}.
We see that FACTS finds {high-performance} ST-blocks in less than 10 minutes on all CTS forecasting tasks, which is negligible compared to the baselines. This indicates that FACTS is efficient enough for deployment in practice.

\begin{table}[!htbp]
    \small
    \centering
    
    \caption{Searching time in GPU hours (h) or minutes (m).}
    \resizebox{0.8\linewidth}{!}{
    \begin{tabular}{c|ccccc}
        \hline
        \multirow{1}{*}{dataset} & AutoSTG & AutoCTS & AutoCTS+ & SimpleSTG & FACTS \\
        \hline
        PEMS-BAY & 278.1 h & 405.7 h & 88.6 h &324.5 h &\textbf{9.2 m} \\
        Electricity & 144.2 h & 197.6 h & 43.8 h &159.4 h & \textbf{7.9 m} \\
        PEMSD7(M) & 81.2 h & 101.5 h & 22.2 h &81.4 h & \textbf{6.5 m} \\
        NYC-TAXI & 13.3 h & 20.2 h & 4.8 h &16.8 h & \textbf{6.0 m} \\
        NYC-BIKE & 12.7 h & 19.8 h & 4.3 h &15.9 h & \textbf{5.7 m} \\
        Los-Loop & 6.8 h & 11.2 h & 2.5 h &7.6 h & \textbf{5.5 m} \\
        SZ-TAXI & 6.6 h & 10.4 h & 2.5 h &7.5 h & \textbf{5.4 m} \\
        \hline
    \end{tabular}
    }
    
    \label{tab: time1}
\end{table}

\noindent\textbf{Ablation studies on the TFL module.} We conduct ablation studies on the TFL module of TAP. We compare FACTS with the following three variants.

\begin{itemize}[leftmargin=*]
    \item[$\bullet$] \textbf{w/o task features:} removes the TFL module from TAP and keeps the rest unchanged.
    \item[$\bullet$] \textbf{w/o semantic features:} removes the semantic feature extraction module from TAP and keeps the rest unchanged.
    \item[$\bullet$] \textbf{w/o statistical features:} removes the statistical feature extraction module from TAP and keeps the rest unchanged.
\end{itemize}

We first randomly sample 100 ST-blocks from the general search space and train them to obtain validation accuracies. We then use the pretrained TAP from each variant and FACTS to predict these ST-blocks respectively to obtain the prediction accuracies.
We use MAE and Spearman's rank correlation coefficient ($\rho$) to quantify the difference between the true validation accuracy and the prediction accuracy of the ST-blocks.
We show results on three unseen datasets under the $P$-24/$Q$-24 forecasting setting in Table~\ref{tab: taskfeat}.

\begin{table*}[!htbp]

 \begin{minipage}[!htbp]{0.30\linewidth}
 
     \caption{Ablation studies on the TFL.}\label{tab: taskfeat}
     \centering
    \begin{adjustbox}{width=\linewidth}
    \begin{tabular}{cc|ccc}
            \hline
            variant &metric & PEMS-BAY & NYC-TAXI & Electricity \\
            \hline
            \multirow{2}{*}{w/o task} & MAE &0.1470 &0.1108 &0.0903\\
             & Spear &0.6401 &0.6671 &0.5915\\
            \multirow{2}{*}{w/o sem} & MAE &0.0851 &0.0703 &0.0652 \\
             & Spear &0.7561 &0.7255 &0.6900\\
            \multirow{2}{*}{w/o stat} & MAE &0.0688 &0.0581 &0.0593 \\
             & Spear &0.7641 &0.7581 &0.7290\\      
             \multirow{2}{*}{FACTS} & MAE &\textbf{0.0410} &\textbf{0.0328} &\textbf{0.0391}\\
             & Spear &\textbf{0.8414} &\textbf{0.8285} &\textbf{0.8151}\\       
             \hline
        \end{tabular}
        
    \end{adjustbox}
  \end{minipage}
  \hspace{0.02\textwidth}
  \begin{minipage}[!htbp]{0.61\linewidth}
  
        \caption{Training time comparison in GPU minutes.}\label{tab: time2}
        \centering
        \begin{adjustbox}{width=\linewidth}
         \begin{tabular}{c|cccccccc}
        \hline
        \multirow{1}{*}{task} & MTGNN & AGCRN & PDFormer & AutoCTS & AutoCTS+ &SimpleSTG & w/o adapt & FACTS \\
        \hline
        PEMS-BAY & 298.6 & 240.5 &252.8 & 325.5 & 283.7 & 280.2 &295.1 &\textbf{110.3} \\
        Electricity & 240.6 & 193.8 & 261.2 & 273.5 &255.7 &260.0  &271.2  &\textbf{94.9} \\
        PEMSD7(M) & 46.3 & 39.1 & 50.4 & 62.4 & 53.3 & 51.7 &47.3 &\textbf{16.4} \\
        NYC-TAXI & 29.2 & 22.6 & 29.6 & 34.1 &23.5 &27.6 &25.4 &\textbf{9.7} \\
        NYC-BIKE & 27.2 & 19.9 & 29.0 & 31.2 &21.0 &26.2 &23.7 &\textbf{8.3} \\
        Los-Loop & 19.6 & 12.8 & 17.5 & 22.7 & 17.8 & 20.4 &18.5 &\textbf{6.3} \\
        SZ-TAXI & 19.3 & 11.9 & 17.1 & 22.9 &30.2 &31.0 &15.8 &\textbf{5.4} \\
        \hline
    \end{tabular}
    
    \end{adjustbox}
   \end{minipage}
\end{table*}

We observe that disabling the TFL module significantly reduces the accuracy of the pretrained TAP on unseen tasks, indicating that TFL is crucial for enabling TAP to generalize to unseen tasks. Next, enabling either the semantic feature extraction module or the statistical feature extraction module can improve the accuracy of TAP, and enabling both achieves the highest accuracy. This shows that they both help identify CTS forecasting tasks and are functionally complementary, so using both results in the best performance.

\noindent\textbf{{The trade-off between pretraining time and accuracy.}}

We study the trade-off between pretraining time and accuracy by varying the number of $combs$ $c$ sampled in each iteration.
Specifically, we construct three variants of FACTS with $c$ equal to $60$, $80$, and $120$ and keep other settings as in FACTS. FACT uses $c=100$.

%
The results under the P-12/Q-12 forecasting setting are shown in Table~\ref{tab: pretrain-12}. We see that the final accuracy generally improves gradually with an increase of the pretraining time, showing that collecting more $combs$ and ST-blocks to pretrain TAP can bring higher zero-shot search performance.
However, when $c$ exceeds $100$, the performance improvement is very limited, indicating that using $c=100$ strikes a good balance between pretraining efficiency and final accuracy.

\begin{table}[!htbp]
    \small
    \centering
    \caption{{Trade-off between pretraining time and accuracy, P-12/Q-12 forecasting.}} 
    \resizebox{0.80\linewidth}{!}{{
    \begin{tabular}{c|c|cccc}
        \hline
        Data & Metric & \makecell{ c=120 \\\textbf{(208.9h)}} & \makecell{ c=100 \\ \textbf{(170.4h)}} & \makecell{ c=80 \\ \textbf{(139.8h)}}& \makecell{ c=60 \\ \textbf{(107.1h)}} \\
        \hline
        & MAE & \textbf{1.509} & \underline{1.517} & 1.584 & 1.654 \\
        PEMS-BAY & RMSE & \textbf{3.287} &\underline{3.292} & 3.488 & 3.891 \\
        & MAPE & \textbf{3.488\%} & \underline{3.494\%} & 3.850\% & 4.491\% \\
        \hline
        & MAE & \textbf{219.951} & \underline{225.814} & 259.963 & 294.065 \\
        Electricity & RMSE & \textbf{1938.105} & \underline{1943.750} & 2215.050 & 2494.129  \\
        & MAPE & \textbf{15.852\%} & \underline{15.871\%} & 18.668\% & 22.913\%\\
        \hline
        & MAE & \textbf{2.542} & \underline{2.551} & 2.685 & 2.778\\
        PEMSD7M & RMSE & \textbf{4.858}& \underline{4.863} & 5.022 & 5.481 \\
        & MAPE & \textbf{6.296\%} & \underline{6.301\%} & 6.645\% & 6.955\% \\
        \hline
        & MAE & \textbf{5.327} & \underline{5.334} & 5.582 & 5.690\\
        NYC-TAXI & RMSE & \textbf{9.460} & \underline{9.472} & 9.785 & 10.302\\
        & MAPE & \underline{37.788\%} & \textbf{37.780\%} & 39.918\% & 41.051\%\\
        \hline
        & MAE & \textbf{1.774} & \underline{1.796} & 1.953 & 2.079 \\
        NYC-BIKE & RMSE & \textbf{2.667} & \underline{2.732} & 3.003 & 3.184\\
        & MAPE & \textbf{49.681\%} & \underline{49.951\%} & 51.391\% & 54.095\%\\
        \hline
        & MAE & \textbf{3.563} & \underline{3.578} & 4.125 & 4.195 \\
        Los-Loop & RMSE & \textbf{6.829} & \underline{6.841} & 7.185 & 7.436\\
        & MAPE &\textbf{9.993\%} & \underline{10.006\%} & 10.496\% & 10.920\% \\
        \hline
        & MAE & \textbf{3.168} & \underline{3.178} & 3.219 & 3.552 \\
        SZ-TAXI & RMSE & \textbf{4.201} & \underline{4.217} & 4.294 & 4.785 \\
        \hline
    \end{tabular}
    }
    }
    \label{tab: pretrain-12}
\end{table}

\subsubsection{Effectiveness of Fast Parameter Adaptation}
\noindent
\\\textbf{Training time comparison.} We first compare the training time of the identified ST-blocks and and those of the baselines on seven unseen tasks. We also consider a variant of FACTS without fast parameter adaptation.
Table~\ref{tab: time2} shows that FACTS takes the least time to train the identified ST-blocks, while still achieving the highest accuracy (see Tables~\ref{tab: 12-12} and~\ref{tab: 168-3}).
In particular, FACTS reduces training time by 62\% to 66\% compared to the variant w/o adapt on the seven unseen tasks, which is evidence of the effectiveness of the proposed fast parameter adaptation strategy.
Furthermore, the total running time of FACTS, i.e., the sum of the search and training time (see Table~\ref{tab: time1}) is lower than the training times of the baselines, which indicates that FACTS is efficient enough for deployment.

\begin{figure}[!htbp]
  \centering

    \includegraphics[width=0.7\linewidth, trim=70 70 70 70,clip]{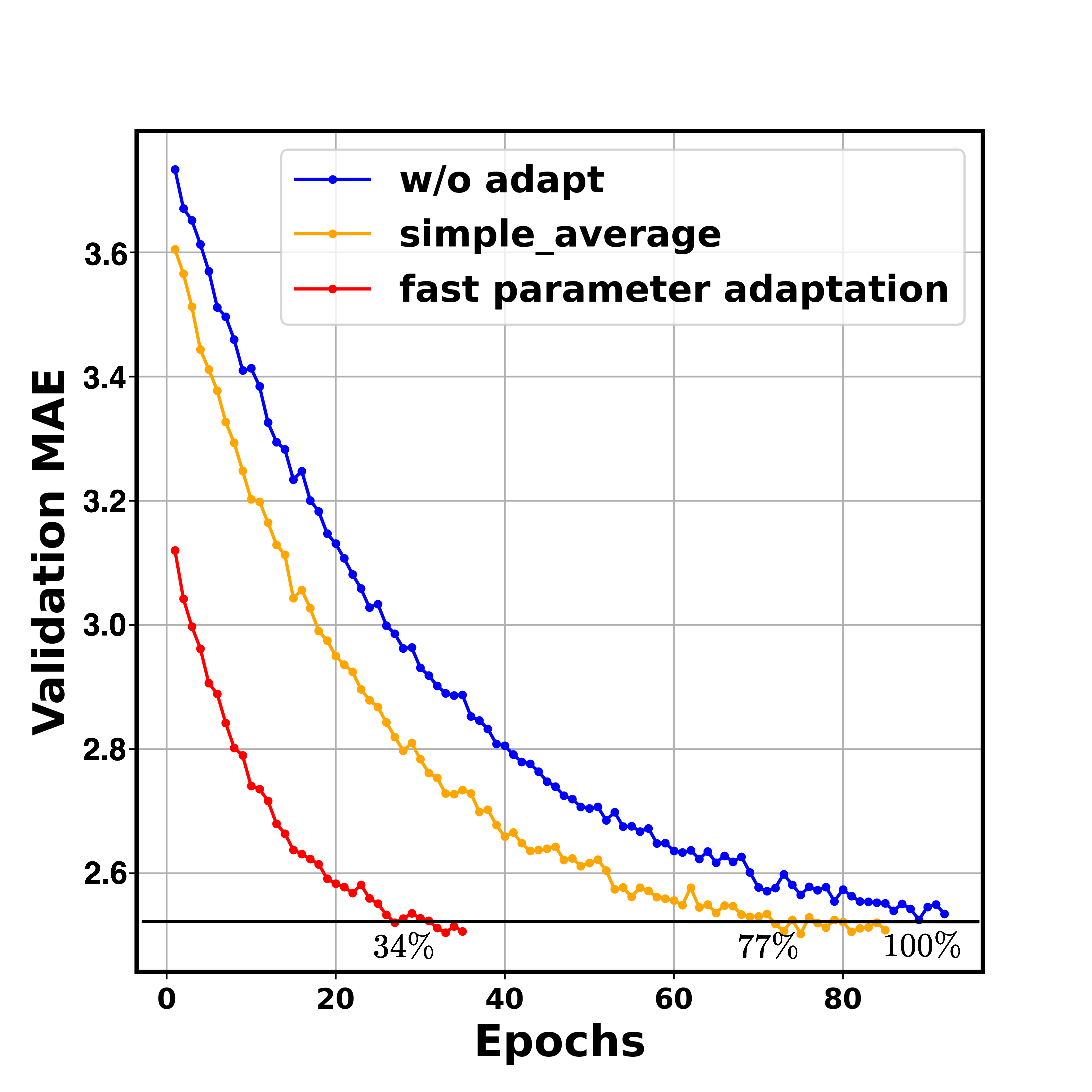} 

  \caption{Efficiency comparison of training strategies. 
  }
  \label{fig: adapt}
\end{figure}

\noindent\textbf{Ablation studies on fast parameter adaptation.} We compare the trend of the validation loss across increasing training epochs for FACTS and two variants on PEMSD7(M).
For the variant w/o adapt, we disable fast parameter adaptation and train the identified ST-block from scratch. For the variant simple\_average, we use simple averaging instead of fast parameter adaptation to inherit the weights from pretrained ST-blocks.
Figure~\ref{fig: adapt} shows that the proposed fast parameter adaptation strategy contributes to the fast convergence of the ST-block without reducing the accuracy, thus indicating the proposed fast parameter adaptation strategy is effective. Next, the variant simple\_average converges faster than training from scratch, but significantly more slowly than FACTS, indicating that the simple parameter inheritance strategy is not as effective as the proposed fast parameter adaptation strategy.

\subsubsection{{Case Study}}
\begin{figure}[!htbp]
\center
\subfigure[Electricity, $P$-12/$Q$-12]{
\begin{minipage}[b]{0.7\linewidth} 
\centering
\includegraphics[width=\linewidth]{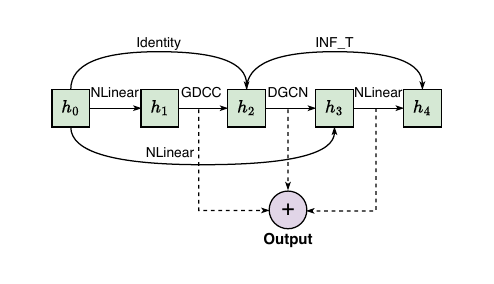}
\label{fig: Electricity-12}
\end{minipage}
}
\subfigure[PEMS-BAY, $P$-24/$Q$-24]{
\begin{minipage}[b]{0.7\linewidth} 
\centering
\includegraphics[width=\linewidth]{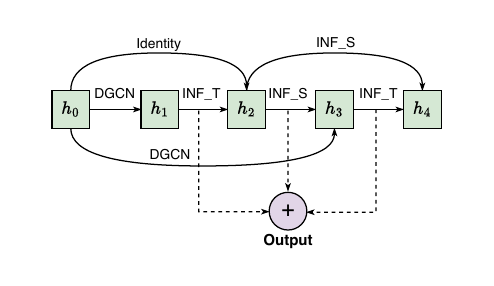}
\label{fig: pems-bay-24}
\end{minipage}
}
\caption{Case study}
\label{fig: case stduy}
\end{figure}

{We show the ST-blocks identified on Electricity and PEMS-BAY in Figure~\ref{fig: case stduy}.
We can see that there are notable differences in the architectures of the ST-blocks.
In particular, the ST-block identified on the Electricity dataset contains three NLinear, which is because Electricity exhibits strong periodicity and weak correlation between time series and because NLinear is good at capturing such patterns~\cite{zeng2023transformers}.
For comparison, the ST-block identified on the PEMS-BAY dataset contains four S-operators, i.e., two DGCN and two Spatial-Informer (INF\_S). As PEMS-BAY is a traffic speed dataset with complex spatial correlations between time series, more S-operators are needed to capture the spatial correlations.}

\section{Related Work}
\subsection{Manual CTS Forecasting}
Manually designed CTS forecasting methods~\cite{lai2018modeling,shih2019temporal,DBLP:conf/ijcai/WuPLJZ19,bai2020adaptive,wu2020connecting,razvanicde2021,MileTS,wang2022multivariate,wang2020multi,DBLP:conf/ijcai/CirsteaG0KDP22,DBLP:conf/icde/CirsteaYGKP22,DBLP:journals/tkde/JinZLCYP23} choose or design S/T operators, assemble them into ST-blocks, and then train the ST-blocks from scratch to make forecasts.
MTGNN~\cite{wu2020connecting} stacks mix-hop GCNs and inception convolution operators sequentially to build an ST-block.
AGCRN~\cite{bai2020adaptive} replace the MLP layers in GRUs by GCNs to build an ST-block.
PDFormer~\cite{jiang2023pdformer} combines semantic, geographic, and temporal self-attention operators in parallel to build an ST-block.
We collect S/T operators and summarize connection rules from these proposals.

\subsection{Automated CTS Forecasting}
Existing automated methods manually design a search space. AutoST (a)~\cite{li2020autost} and AutoSTG~\cite{pan2021autostg} empirically select a few S/T operators to construct a small search space.
AutoCTS~\cite{wu2022autocts} compares the accuracy of S/T operators on several CTS forecasting datasets and selects a few S/T operators with the highest accuracy to construct a search space.
AutoST (b)~\cite{li2022autost} manually designs three S/T operators that consider the order between spatial and temporal modules to construct a small search space.
SimpleSTG~\cite{xu2022understanding} proposes three pruning strategies, and for each strategy, it manually removes poor design choices based on the accuracy of sampled ST-blocks.
Considering search efficiency, AutoST (a), AutoSTG, AutoCTS, and AutoST (b) employ a gradient-based search strategy, which requires training a supernet.
SimpleSTG employs random search to find the optimal ST-block from a set of randomly sampled ST-blocks through fully training.
AutoCTS+~\cite{wu2023autocts+} proposes a comparator to compare the accuracy of different ST-blocks, which necessitates the training of many ST-blocks to obtain their validation accuracies.
Considering training efficiency, existing automated methods train identified ST-blocks from scratch.

\section{Conclusion}

We propose FACTS, an efficient and fully automated CTS forecasting framework that can find an ST-block that offers {high-performance} accuracy on arbitrary unseen tasks and can make forecasts in minutes.
Experimental results on seven commonly used CTS forecasting datasets show that the FACTS framework is capable of state-of-the-art accuracy and efficiency.
%
{FACTS includes several hyperparameters, and methods exist that may reduce these hyperparameters, such as reinforcement learning.
However, introducing the black-box optimization algorithm may also bring limitations to FACTS, such as introducing additional complexity and requiring more computing resources and time. This thus requires further research and we leave it as future work.}


\clearpage

\bibliographystyle{ACM-Reference-Format}
\bibliography{sample}

\end{document}